\documentclass[]{beingbeyond}
\usepackage{enumitem}
\usepackage[toc,page,header]{appendix}

\usepackage{amsmath,amssymb,amsthm,bm}
\usepackage{colortbl}
\usepackage{algorithm}
\usepackage{algpseudocode}
\usepackage{titletoc}
\usepackage{wrapfig}
\usepackage{makecell}
\usepackage{adjustbox}
\usepackage{tabularx}
\usepackage{bbm}

\definecolor{bestgreen}{HTML}{D5F5E3}
\definecolor{secondblue}{HTML}{D6EAF8}
\definecolor{deltared}{HTML}{FADBD8}
\definecolor{deltagreen}{HTML}{D5F5E3}

\newcommand{\cD}{\mathcal{D}}

\newcommand{\KL}{\mathrm{KL}}
\newcommand{\Unif}{\mathrm{Unif}}
\newcommand{\sg}[1]{\mathrm{sg}\!\left(#1\right)}

\newcommand{\E}{\mathbb{E}}
\newcommand{\R}{\mathbb{R}}

\newcommand{\eg}{\textit{e.g.}}

\usepackage{coloredtheorem}
\cthnewtheorem{proposition}{Proposition}

\theoremstyle{definition}

\theoremstyle{remark}

\crefname{equation}{Eq}{Eqs.}
\Crefname{equation}{Eq}{Eqs.}
\crefname{theorem}{Theorem}{Theorems}
\Crefname{theorem}{Theorem}{Theorems}
\crefname{proposition}{Proposition}{Propositions}
\Crefname{proposition}{Proposition}{Propositions}
\crefname{section}{Section}{Sections}
\Crefname{section}{Section}{Sections}
\crefname{subsection}{Section}{Sections}
\Crefname{subsection}{Section}{Sections}
\crefname{algorithm}{Algorithm}{Algorithms}
\Crefname{algorithm}{Algorithm}{Algorithms}

\title{Conservative Offline Robot Policy Learning via Posterior-Transition Reweighting}

\author{{\bfseries
Wanpeng Zhang$^{1,3}$~
Hao Luo$^{1,3}$~
Sipeng Zheng$^{3}$~
Yicheng Feng$^{1,3}$~
Haiweng Xu$^{1,3}$ \\
Ziheng Xi$^{2,3}$~
Chaoyi Xu$^{1,3}$~
Haoqi Yuan$^{1,3}$~
Zongqing Lu$^{1,3,\dagger}$
}}

\affiliation{
$^{1}$Peking University \quad
$^{2}$Tsinghua University \quad
$^{3}$BeingBeyond \quad
}

\webpage{\url{https://research.beingbeyond.com/ptr}}

\abstract{
Offline post-training adapts a pretrained robot policy to a target dataset by supervised regression on recorded actions. In practice, robot datasets are heterogeneous: they mix embodiments, camera setups, and demonstrations of varying quality, so many trajectories reflect recovery behavior, inconsistent operator skill, or weakly informative supervision. Uniform post-training gives equal credit to all samples and can therefore average over conflicting or low-attribution data. We propose \textbf{Posterior-Transition Reweighting (PTR)}, a reward-free and conservative post-training method that decides how much each training sample should influence the supervised update. For each sample, PTR encodes the observed post-action consequence as a latent target, inserts it into a candidate pool of mismatched targets, and uses a separate transition scorer to estimate a softmax identification posterior over target indices. The posterior-to-uniform ratio defines the PTR score, which is converted into a clipped-and-mixed weight and applied to the original action objective through self-normalized weighted regression. This construction requires no tractable policy likelihood and is compatible with both diffusion and flow-matching action heads. Rather than uniformly trusting all recorded supervision, PTR reallocates credit according to how attributable each sample’s post-action consequence is under the current representation, improving conservative offline adaptation to heterogeneous robot data.
}

\checkdata[Date]{Mar 17, 2026}

\begin{document}

\maketitle

\begingroup
\renewcommand\thefootnote{\fnsymbol{footnote}} 
\setcounter{footnote}{0}
\footnotetext[2]{Correspondence to Zongqing Lu $<$lu@beingbeyond.com$>$.}
\endgroup


\section{Introduction}

Pretrained vision-language-action (VLA) policies~\citep{brohan2022rt,zitkovich2023rt,kim2024openvla} provide a practical foundation for robot learning.
Large-scale pretraining~\citep{black2024pi_0,intelligence2025pi_,bjorck2025gr00t,luo2025being,luo2026being} encodes broad robot priors into a shared backbone, and supervised post-training adapts the policy to a target setting.
This pipeline stays purely offline and keeps deployment simple.

Data heterogeneity is the core challenge.
Large robot collections mix trajectories from different embodiments, camera viewpoints, control delays, and diverse teleoperators~\citep{khazatsky2024droid,collaboration2023open,walke2023bridgedata}.
Even within one embodiment, operator skill varies: some demonstrations are near-optimal, while others contain recovery behaviors or hesitations.
Across embodiments, similar images can correspond to different kinematic solutions.
Logged action chunks are therefore multi-modal, with uneven quality and substantial suboptimal supervision.

Cross-embodiment mixtures also carry a latent positive-transfer potential.
Different robots can demonstrate the same high-level skill and provide additional coverage of task-relevant progress, even when their low-level action chunks differ.
Recent VLAs such as Being-H0.5~\citep{luo2026being} enable this by mapping heterogeneous robots into a unified action space.
The difficulty is to exploit the signal selectively without incurring negative transfer from embodiment-specific artifacts.

This paper proposes a simple idea: use observed post-action consequences as a reward-free signal for deciding which recorded chunks deserve more credit.
Offline datasets record not only an action chunk but also what happens after it.
PTR turns this observation into an identification test.
Given the current policy representation and the recorded chunk, can the matched post-action consequence be identified among a pool of mismatched alternatives?
Concentrated identification posteriors indicate attributable, high-quality chunks that receive more weight.
Diffuse posteriors indicate ambiguous or suboptimal samples that are down-weighted.
Conservative clipping and mixture constraints keep the induced distribution shift bounded.

When demonstrations are already consistent, PTR weights stay close to uniform and the method reduces to standard post-training.
The gains come from reallocating credit along two axes:
PTR raises the performance floor by suppressing suboptimal and conflicting supervision, and it raises the ceiling by selectively leveraging cross-embodiment coverage when post-action consequences align across sources.

Our contributions are threefold:
\begin{itemize}
    \item A reward-free sample scoring mechanism that converts post-action consequences into an identification posterior, whose log ratio to the uniform baseline measures how attributable each recorded chunk is to the current policy context.
    \item A conservative weight mapping that bounds the induced distribution shift while preserving the original supervised action objective, with formal guarantees connecting the score to KL divergence and the weight to bounded density ratios.
    \item Empirical validation on simulation benchmarks and $12$ real-robot tasks across three embodiments, demonstrating the general effectiveness of PTR.
\end{itemize}


\section{Related Work}

\textbf{Vision-language-action models.}
Vision-language-action (VLA) models unify vision encoders~\citep{radford2021learning,feng2025videoorion,luo2025openmmego}, language models~\citep{zhang2025beingvl0,zhang2025unified}, and action decoders into end-to-end robot policies~\citep{luo2026jointalignedlatentaction,feng2026spatialaware}.
Early systems such as RT-1~\citep{brohan2022rt} and RT-2~\citep{zitkovich2023rt} demonstrated that transformer-based architectures can learn generalizable robot control from large datasets.
PaLM-E~\citep{driess2023palm} showed that multimodal language models can ground in embodied tasks.
Open-source generalist policies including OpenVLA~\citep{kim2024openvla} and Octo~\citep{team2024octo} have made VLA pretraining broadly accessible.
Autoregressive VLAs tokenize actions and predict them sequentially~\citep{kim2024openvla,pertsch2025fast}, while a growing family of models generates action chunks via continuous generative processes.
$\pi_0$~\citep{black2024pi_0} and $\pi_{0.5}$~\citep{intelligence2025pi_} use flow matching~\citep{lipman2022flow,liu2022flow} as the action head.
GR00T N1~\citep{bjorck2025gr00t} adopts a dual-system architecture with a DiT-based action generator.
Diffusion Policy~\citep{chi2025diffusion} applies denoising diffusion~\citep{ho2020denoising} to visuomotor control.
Being-H0.5~\citep{luo2026being} combines a Mixture-of-Transformers backbone with flow matching and introduces a unified action space that maps heterogeneous robots to shared semantic slots, enabling cross-embodiment pretraining.
Large-scale cross-embodiment datasets~\citep{collaboration2023open,khazatsky2024droid,walke2023bridgedata} provide the data substrate for these models but also introduce the heterogeneity and suboptimal demonstrations that motivate PTR.
PTR operates at the post-training stage of such systems and is compatible with both autoregressive and generative action heads.

\textbf{Offline policy improvement and data reweighting.}
Standard behavioral cloning treats all demonstrations equally.
Dataset composition and demonstration quality significantly affect imitation learning performance~\citep{belkhale2023data,hejna2024re,wang2026rethinking}.
Weakly supervised quality estimators~\citep{kuhar2023learning}, representation modulation~\citep{zhang2025dig}, and mutual-information-based data curation~\citep{hejna2025robot} attempt to address this.
A classical alternative is advantage-weighted regression (AWR)~\citep{peng2019advantage}, which casts policy improvement as supervised learning with exponential weights $\exp(A/\beta)$.
Reward-weighted regression~\citep{peters2007reinforcement}, REPS~\citep{peters2010relative}, and MPO~\citep{abdolmaleki2018maximum} share this exponential-weight structure.
Reward-conditioned policies~\citep{kumar2019reward} and Decision Transformer~\citep{chen2021decision} condition on returns rather than reweighting.
PTR adopts the same exponential weight form as AWR but replaces reward-based advantages with a reward-free identification score derived from post-action consequences.
A growing line of work~\citep{li2025simplevla,guo2025improving,huang2025co} applies reinforcement learning to VLA fine-tuning~\citep{chen2025pirl,lu2025vla,liu2025can}.
These methods require reward signals or online interaction~\citep{zhai2025vision,zhang2024grape,frans2025diffusion,chen2025conrft,zhang2025robustvla}.
PTR uses no reward, no value function, and no policy gradient; its connection to this literature is structural (the exponential weight form from KL-regularized optimization~\citep{levine2018reinforcement}) rather than algorithmic.
The identification posterior builds on InfoNCE~\citep{gutmann2010noise,oord2018representation} and causality assignment methods~\citep{huang2020causal,zhang2024tackling}.
The conservative constraints (clipping, mixture, self-normalization) mirror truncated importance weighting~\citep{schulman2017proximal,espeholt2018impala} and self-normalized estimators~\citep{swaminathan2015self} from the off-policy and offline RL literature~\citep{levine2020offline,fujimoto2019off,kumar2019stabilizing}.
PTR adapts these principles to reward-free supervised post-training with an identification-based score.


\section{Preliminaries and Notation}
\label{sec:prelim}

\textbf{Robot dataset and training tuples.}\par
Each sample from an offline dataset $\cD$ is a five-tuple $(o_t, s_t, l, a_{t:t+L-1}, o_{t+\Delta})$.
It contains visual observation $o_t$, state $s_t\in\R^{d_s}$, instruction $l$, action chunk $a_{t:t+L-1}\in\R^{L\times d_a}$, and future observation $o_{t+\Delta}$.
Here $L$ is fixed, while $\Delta$ may vary across samples. Only $(o_t,s_t,l)$ is used at inference; $o_{t+\Delta}$ serves exclusively as a training-time target for the identification test.

\textbf{VLA backbone and unified action space.}
Let $f_\phi$ denote a transformer backbone that maps $(o_t,l)$ to hidden states $H_t$ and a pooled context $h_t$.
Being-H0.5~\citep{luo2026being} maps heterogeneous robots into a shared $200$-dimensional action space with sparse semantic slot assignments, so that similar motor components always occupy the same dimensions regardless of embodiment.
PTR inherits this representation.

\textbf{Action heads and post-training objective.}
The action head maps $(h_t,s_t)$ to $\hat a_{t:t+L-1}\in\R^{L\times d_a}$.
For flow-matching heads~\citep{lipman2022flow,liu2022flow}, the per-sample loss is
\begin{equation}
\label{eq:fm_loss}
\ell_{\mathrm{act}}(\phi; h_t, s_t, a_{t:t+L-1})
= \bigl\| v_\phi\bigl(\sigma a_{t:t+L-1} + (1{-}\sigma)\epsilon,\;\sigma,\;h_t,\;s_t\bigr) - (a_{t:t+L-1} - \epsilon) \bigr\|^2,
\end{equation}
where $\epsilon\sim\mathcal N(0,I)$ and $\sigma\sim p(\sigma)$.
Diffusion heads~\citep{chi2025diffusion,ho2020denoising} admit a similar form.
Uniform post-training minimizes
\begin{equation}
\label{eq:uniform_sft}
\min_\phi\;\E_{(o_t,s_t,l,a_{t:t+L-1})\sim\cD}\bigl[\ell_{\mathrm{act}}(\phi;h_t,s_t,a_{t:t+L-1})\bigr].
\end{equation}

\begin{figure}[t]
    \centering
    \includegraphics[width=\linewidth]{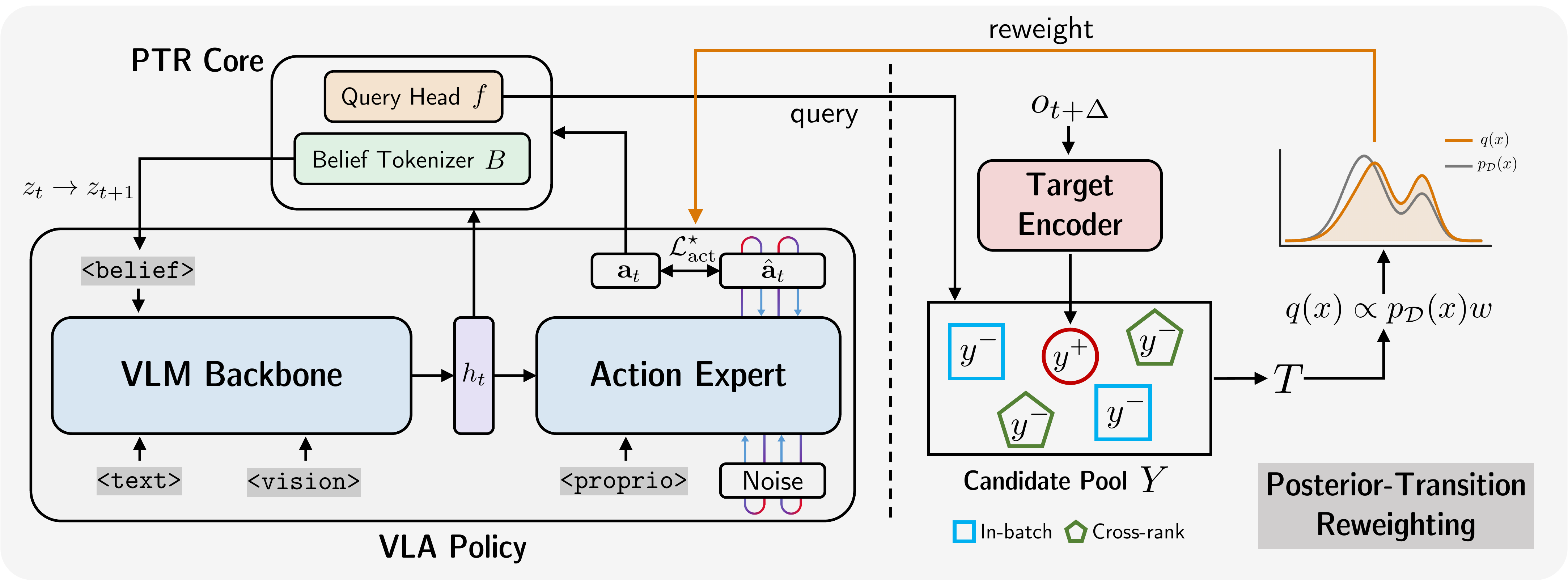}
    \caption{Overview of PTR.
    Left: the standard policy stack (backbone + action expert) is augmented with a lightweight scorer and a BeliefTokenizer.
    Right: for each training chunk, the scorer identifies the matched post-action target among mismatched candidates; the resulting identification posterior is converted into a conservative weight that rescales the supervised action loss.
    No reward labels or policy likelihoods are needed.}
    \label{fig:framework}
\end{figure}


\section{Posterior-Transition Reweighting}
\label{sec:ptr}

PTR overlays standard offline post-training with a conservative reweighting mechanism.
A lightweight consequence encoder and transition scorer produce per-sample weights from observed post-action consequences, without requiring reward labels or a tractable policy likelihood.
The section is organized along the data flow: we first describe the belief proxy tokens that summarize interaction history (\cref{sec:belief_tokenizer}), then the identification scorer that converts post-action consequences into a per-sample quality signal (\cref{sec:ptr_score}), followed by the theoretical foundations that justify reading this signal as a density ratio and KL divergence (\cref{sec:theory}).
We then present the conservative weight mapping that bounds distribution shift (\cref{sec:ptr_reweight}), the adaptive controller that keeps the scorer in a stable operating range (\cref{sec:adarefiner}), and the practical training pipeline (\cref{sec:ptr_pipeline}).

\subsection{BeliefTokenizer}
\label{sec:belief_tokenizer}

PTR maintains $M$ compact belief proxy tokens $z_t\in\R^{M\times d}$ that are appended to the backbone input.
These tokens summarize pre-action interaction history and help define what counts as a similar context under partial observability.
For a segment starting at $t=t_0$, the initial tokens $z_{t_0}=z_{\mathrm{init}}$ are learned.
At each chunk index $t \ge t_0$, the forward pass produces:
\begin{align}
\label{eq:contract}
(H_t, h_t) &= f_\phi(o_t, l, z_t), \\
E_t &= g_\phi(h_t, s_t, a_{t:t+L-1})\in\R^{L\times d}, \qquad e_t=\mathrm{pool}(E_t), \\
z_{t+1} &= \sg{B(H_t, E_t)}.
\end{align}
Here $H_t$ denotes token-level backbone hidden states, $h_t$ is a pooled context representation, $E_t$ is the sequence of action-channel tokens, and $e_t$ is its pooled summary used by the scorer.
The BeliefTokenizer $B$ compresses current-step features into next-step tokens via soft causal assignments.
The stop-gradient on $z_{t+1}$ blocks gradients through time; the tokenizer learns from current-step losses only.
An adaptive scale controller monitors identification statistics and adjusts the scorer temperature $\tau_{\mathrm{score}}$, the advantage scaling $\beta$, and the hard-negative ratio within fixed bounds to keep training stable (Section~\ref{sec:adarefiner}).

\textbf{Soft causal tokenization.}
For a chunk $t$, let $C_t\in\R^{L\times d}$ denote per-step context features and $A_t\in\R^{L\times d}$ the corresponding action-channel features.
In code, $C_t$ corresponds to transformer hidden states on the action-token positions and $A_t$ to the action embeddings used by the action head; $d$ is the hidden size of that action-channel representation.
The tokenizer compresses these $L$ per-step features into $M$ belief proxy tokens ($L=16$, $M=4$). It first fuses the two streams:
\begin{equation}
\label{eq:tok_fuse}
c_{t,i}=\tanh\!\bigl(W_f [C_{t,i};A_{t,i}]\bigr)\in\R^d,\qquad i=1,\ldots,L,
\end{equation}
then computes assignment logits for $M$ slots, normalized over time:
\begin{equation}
\label{eq:tok_assign}
\xi_{t,i,m}=(W_a c_{t,i})_m,\qquad
\pi_{t,i,m}=\frac{\exp(\xi_{t,i,m}/\tau_{\mathrm{tok}})}{\sum_{j=1}^{L}\exp(\xi_{t,j,m}/\tau_{\mathrm{tok}})}.
\end{equation}
The merged belief tokens are weighted averages:
\begin{equation}
\label{eq:tok_merge}
z_{t+1,m}=\sum_{i=1}^{L}\pi_{t,i,m}\,c_{t,i},\qquad m=1,\ldots,M.
\end{equation}
We also reconstruct per-step features as
\begin{equation}
\label{eq:tok_perstep}
\tilde c_{t,i}=\sum_{m=1}^{M}\pi_{t,i,m}\,z_{t+1,m}.
\end{equation}
The recursion in \cref{eq:contract} passes $z_{t+1}$ to the next chunk with stop-gradient.

\textbf{Tokenizer regularizers.}
Two auxiliary losses prevent degenerate tokenizer behavior.
An entropy term encourages each slot to attend decisively rather than spreading weight uniformly.
Let $\Pi_t\in\R^{L\times M}$ stack $\pi_{t,i,m}$.
The average entropy of each slot's distribution over time is
\begin{equation}
\label{eq:tok_entropy}
H_{\mathrm{tok}}
=
-\frac{1}{M}\sum_{m=1}^M \sum_{i=1}^{L}\pi_{t,i,m}\log(\pi_{t,i,m}).
\end{equation}
Adding $H_{\mathrm{tok}}$ to the loss with a positive coefficient encourages the tokenizer to form more decisive groupings.
To prevent collapse where multiple slots attend to the same subset of time steps, we penalize the slot Gram matrix:
\begin{equation}
\label{eq:tok_div}
D_{\mathrm{tok}}
=
\left\|\Pi_t^\top \Pi_t - I\right\|_F^2.
\end{equation}
The combined tokenizer loss is $\mathcal L_{\mathrm{tok}}=\lambda_{\mathrm{ent}}H_{\mathrm{tok}}+\lambda_{\mathrm{div}}D_{\mathrm{tok}}$.

\subsection{Posterior transition score}
\label{sec:ptr_score}

With belief proxy tokens providing a compact summary of history, PTR builds a reward-free quality signal from an identification posterior over post-action consequences.
The \emph{posterior} here refers to a softmax distribution over candidate targets in a finite pool, not a Bayesian posterior or a predictive dynamics model.

\textbf{Post-action targets.}
PTR encodes the observed post-action observation into the matched latent target $y_{t,0}^{+}=\sg{g(o_{t+\Delta})}$,
where $g$ is a momentum (EMA) target encoder (distinct from the action-channel head $g_\phi$ in \cref{eq:contract}). The motivation is to trace the causal effect of actions from future consequences~\citep{huang2020causal,zhang2024tackling}.
PTR works in a latent target space rather than raw pixels.
Reweighting only needs a compact representation that makes consequences distinguishable for identification.
We reuse an intermediate layer of the policy's own vision tower and maintain it with EMA, following momentum encoders in contrastive learning~\citep{he2020momentum,grill2020bootstrap}.
A frozen target space becomes misaligned as the policy representation evolves; a fully online target is unstable.
EMA is a stable compromise.

Concretely, the target encoder $g$ extracts features from vision layer $12$ of InternViT-300M-448px and is updated via exponential moving average with decay $\mu=0.999$: $g \leftarrow \mu g + (1-\mu) v$.
All target features are L2-normalized before entering the candidate pool.
Post-action targets are always stop-gradient features; the future observation is never fed back into the action policy as an additional input, keeping PTR in the offline post-training regime.

\textbf{Candidate pool.}
PTR forms an ordered candidate set $Y_t=\{y_{t,0}^{+}\}\cup Y_t^{-}$ with $Y_t^{-}=\{y_{t,1}^{-},\ldots,y_{t,K}^{-}\}$,
where $y_{t,1}^{-},\ldots,y_{t,K}^{-}$ are mismatched targets from other samples.
These are \emph{target replacements}, not trajectory splicing.

For each minibatch, we compute matched targets $y_{t,0}^{+}$ for valid samples and draw mismatched targets from three sources:
(i)~in-batch targets from other samples, (ii)~cross-rank gathered targets from other GPUs, and (iii)~a FIFO queue storing targets from previous iterations.
All targets are treated as constants for the current update; the queue and gather are non-differentiable and exist only to enlarge the candidate pool.
Negatives are formed after removing the current sample's matched target, so the scorer must identify the correct post-action consequence against genuinely mismatched alternatives.
When the refiner increases the hard-negative ratio, harder samples are mixed into the same pool rather than handled by a separate objective.
In the default configuration, the FIFO queue holds $1024$ entries, each minibatch draws up to $64$ queue negatives per sample, and targets are gathered across all $8$ GPUs via a non-differentiable \texttt{all\_gather} before pool construction.

When a chunk lacks a valid post-action observation, we omit it from the scorer-side losses and use the conservative fallback $w_t=1$, so the sample contributes exactly as in uniform post-training.

\textbf{Identification posterior.}
The scorer forms a query embedding $u_t=f(h_t,e_t)$ from the current representation $(h_t,e_t)$ in \cref{eq:contract}, using a lightweight projection head $f$ (distinct from the backbone $f_\phi$).
It computes a cosine-similarity logit against each candidate, $d_{t,i} := \langle \mathrm{norm}(u_t),\,\mathrm{norm}(y_{t,i})\rangle / \tau_{\mathrm{score}}$,
and defines the identification posterior, where $I_t\in\{0,\ldots,K\}$ indexes the candidate believed to be the matched target:
\begin{equation}
\label{eq:posterior}
\hat p(I_t=0\mid h_t,e_t,Y_t)
=
\frac{\exp(d_{t,0})}{\sum_{j=0}^{K}\exp(d_{t,j})}.
\end{equation}
This posterior has the same form as InfoNCE identification objectives~\citep{gutmann2010noise,oord2018representation}.
In our implementation, the action head already computes $E_t$ and we set $e_t=\mathrm{pool}(E_t)$; the figure shows $a_{t:t+L-1}$ for clarity.

The scorer conditions on an explicit action channel.
In code, $e_t$ is a pooled representation of the action-channel tokens already used by the action head in the same forward pass, so PTR does not introduce a second action encoder.
The notation $g_\phi$ in \cref{eq:contract} should be read purely as the action-channel projector used by the action head; it is unrelated to the EMA target encoder $g$.

To prevent the scorer from collapsing into a context-only shortcut, we add an action-sensitivity regularizer.
The projection from the pooled action summary into the scorer's query space is a two-layer MLP with Xavier initialization on both layers.
Let $u_t^+=f(h_t,e_t)$ and $u_t^-=f(h_t,\tilde e_t)$, where $\tilde e_t$ is obtained by permuting action features within the minibatch.
Let $d_{t,0}^{+}$ and $d_{t,0}^{-}$ denote the matched-target logits computed with $u_t^+$ and $u_t^-$ respectively.
The ranking loss is
\begin{equation}
\label{eq:rank}
\mathcal L_{\mathrm{rank}}(\theta)
=
-\E_t\Bigl[\log\sigma\Bigl(d_{t,0}^{+}-d_{t,0}^{-}\Bigr)\Bigr].
\end{equation}

\textbf{PTR score.}
We define the posterior-to-uniform ratio as
\begin{equation}
\label{eq:score}
T_t
\triangleq
\log \frac{\hat p(I_t=0\mid h_t,e_t,Y_t)}{1/|Y_t|}.
\end{equation}
If the posterior is uniform over the candidate pool, then $T_t=0$ and the sample falls back to uniform supervision.
If the posterior is concentrated on the matched target, then $T_t>0$.
Because the score is produced by a separate scorer, PTR does not require the policy itself to expose a likelihood and remains compatible with flow-matching action heads.

\textbf{Natural suppression of suboptimal demonstrations.}
Robot datasets inevitably contain suboptimal trajectories: recovery behaviors, hesitant motions, or demonstrations from less-skilled operators.
For such samples the post-action observation $o_{t+\Delta}$ is often less distinctive under the pre-action context $(h_t, e_t)$, so the consequence becomes harder to attribute to the recorded chunk.
The identification posterior therefore spreads across the candidate pool, yielding a low or negative PTR score: ambiguous samples stay near uniform weight, while clearly counter-evidential ones are down-weighted.
In contrast, high-quality demonstrations produce distinctive post-action consequences that concentrate the posterior, resulting in $T_t \gg 0$ and higher credit.
This mechanism provides a \emph{floor}: PTR removes extra emphasis from suboptimal data and down-weights it whenever the matched target becomes less likely than the pool-average alternative.

\subsection{Theoretical foundations}
\label{sec:theory}

The PTR score defined above is an empirical quantity computed from a finite candidate pool.
This subsection establishes its theoretical grounding by following the mathematical dependency chain: we first formalize the candidate-set identification model, show that Bayes-optimal logits recover a density ratio (Proposition~\ref{prop:density_ratio}), use this to connect the PTR score to a KL divergence (Proposition~\ref{prop:kl_lens_main}), derive the exponential weight form from a KL-regularized objective, and analyze how tilting reallocates weight across data sources (Proposition~\ref{prop:mixture_main}).
Formal proofs of all three propositions are collected in Appendix~\ref{app:proofs}.

\textbf{Candidate-set model.}
\label{sec:candidate_model}
All theoretical results rest on a common probabilistic model of the identification task.
The model is standard in the contrastive learning literature~\citep{oord2018representation,gutmann2010noise} and is included here to fix notation.

Fix a context representation $(h,e)$ and a baseline target distribution $p_-(y\mid h)=p_N(y\mid h)$.
Assume the positive distribution $p_+(\cdot\mid h,e)$ is absolutely continuous with respect to $p_-(\cdot\mid h)$ on the support induced by the candidate pool, so the density ratio $r(y)=p_+(y\mid h,e)/p_-(y\mid h)$ is well-defined there.
First draw a candidate position $I$ uniformly from $\{0,\ldots,K\}$.
Then sample the matched candidate $Y_I$ from $p_+(y\mid h,e):=p(y\mid h,e)$ and sample every mismatched candidate $Y_j$ ($j\neq I$) independently from $p_-(y\mid h)$.
The ordered training view used by PTR is obtained by conditioning on $I=0$, so the matched target sits at index $0$ and the remaining $K$ entries act as negatives.
The distribution $p_-$ is the population counterpart of the practical pool-construction rule described in Section~\ref{sec:ptr_score}; it can absorb any fixed mixture of in-batch, cross-rank, queued, or harder same-task negatives.

A scorer produces logits $s(h,e,y)$ and induces the identification posterior
\begin{equation}
\label{eq:app_posterior}
\hat p(I=0\mid h,e,Y)
=
\frac{\exp s(h,e,Y_0)}{\sum_{j=0}^{K}\exp s(h,e,Y_j)}.
\end{equation}

\textbf{Density-ratio form of optimal logits.}
The Bayes-optimal scorer recovers a log density ratio between the action-conditioned and baseline target distributions.
This result underpins the KL and entropy interpretations and clarifies why the PTR score can serve as a meaningful quality signal even though it is computed from a finite candidate set.

\begin{cthproposition}{Bayes-optimal logits recover a density ratio}
\label{prop:density_ratio}
Under the candidate-set model above, Bayes-optimal shared per-candidate logits for the identification task in Eq~\eqref{eq:app_posterior} can be written as
\begin{equation}
\label{eq:app_density_ratio}
s^\star(h,e,y)
=
\log\frac{p(y\mid h,e)}{p_N(y\mid h)} + b(h,e),
\end{equation}
where $b(h,e)$ does not depend on $y$.
\end{cthproposition}

This proposition has two practical consequences.
First, the identification scorer is not learning an arbitrary discriminative function: at optimality, the logits recover a principled statistical quantity (the log density ratio) that measures how much the action changes the distribution over future observations.
Second, the additive constant $b(h,e)$ cancels in the softmax posterior of Eq~\eqref{eq:app_posterior}, so the PTR score depends only on the density ratio induced by the chosen baseline pool and not on any candidate-independent offset.

\textbf{KL and entropy views of the PTR score.}
With the density-ratio form in hand, we can relate the population PTR score to a KL divergence.
For fixed $(h,e)$, define $p_+(y)=p(y\mid h,e)$, $p_-(y)=p_N(y\mid h)$, and $r(y)=p_+(y)/p_-(y)$.
By Proposition~\ref{prop:density_ratio}, the Bayes-optimal identification posterior takes the form
\begin{equation}
    p^\star(I=0\mid h,e,Y)=\frac{r(Y_0)}{\sum_{j=0}^{K} r(Y_j)}.
\end{equation}
Under the bounded-ratio regularity condition stated below, the law of large numbers drives the denominator toward its expectation (which equals one under $p_-$), so the score converges pointwise to $\log r(Y_0)$.
Taking expectations over $Y_0\sim p_+$ then recovers $\KL(p_+\|p_-)$, which is the content of the following Proposition~\ref{prop:kl_lens_main}.

\begin{cthproposition}{Large-candidate limit yields a KL score}
\label{prop:kl_lens_main}
Let $p_+(y)=p(y\mid h,e)$ and $p_-(y)=p_N(y\mid h)$ denote the positive (action-conditioned) and negative (baseline, drawn from the candidate pool) target distributions, and let $T^\star$ be the PTR score under the Bayes-optimal identification posterior.
Assume in addition that on the common support there exist constants $0<c\le C<\infty$ such that $c\le r(y)=p_+(y)/p_-(y)\le C$.
As the candidate-set size $K\to\infty$,
\begin{equation}
\label{eq:kl_score_main}
\E\!\left[T^\star \mid h,e\right]\ \longrightarrow\ \KL\bigl(p_+(y)\,\|\,p_-(y)\bigr).
\end{equation}
\end{cthproposition}

Samples whose action makes the next observation highly distinguishable from random alternatives have large $\KL(p_+\|p_-)$ and receive high PTR scores.
Suboptimal or noisy actions blur this distinction, yielding low or negative scores.
Proposition~\ref{prop:kl_lens_main} is a population lens for a fixed baseline $p_-$.
In practice, finite candidate sets, cross-rank reuse, and hard-negative mining replace $p_-$ by an empirical mixture that varies slowly over training.
The equality in Eq~\eqref{eq:kl_score_main} is therefore not asserted sample-by-sample; it explains what quantity the learned score approaches when the pool is large and the sampling rule is stable.

\textbf{How the score behaves in practice.}
Because $T_t$ is centered against the uniform posterior over the realized candidate pool, ambiguous samples concentrate near $0$.
Samples whose matched target is systematically less supported than the pool average can have $T_t<0$ and, after exponentiation and clipping, receive weights below one.
Operationally, PTR performs a \emph{conservative reallocation of credit}: clear samples are amplified, ambiguous samples revert toward uniform, and strongly counter-evidential samples are suppressed.

\textbf{Entropy view under a uniform baseline.}
If targets are indexed by a finite universe $\Omega$ and the target embedding is a lossless encoding of the latent target identity $U$, then, in the same large-candidate limit, choosing a uniform baseline $p_N(U\mid h)=\Unif(\Omega)$ gives
\[
\E[T^\star\mid h,e]\ \longrightarrow\ \log|\Omega|-H(U\mid h,e).
\]
This recovers the entropy interpretation: samples with low conditional entropy (highly predictable consequences) receive high PTR scores.

\textbf{Compatibility with generative action heads.}
The density-ratio and entropy/KL lenses apply to the score estimator (the identification classifier), not to the policy.
They do not require the policy to provide a likelihood over actions, which is why PTR is compatible with diffusion and flow-matching action heads.

\textbf{KL-regularized tilting yields exponential weights.}
PTR's exponential weight $w_t=\exp(T_t/\beta)$ arises as the solution to a KL-regularized score-maximization problem (Eq~\eqref{eq:weight}).
This is a classical result in the policy search literature~\citep{peters2010relative,peters2007reinforcement,abdolmaleki2018maximum}; we reproduce the short derivation here to make the paper self-contained and to clarify the role of the temperature $\beta$.

Consider a base distribution $p_{\cD}(x)$ and a measurable score function $J(x)$ such that the partition function $Z_\beta:=\E_{p_{\cD}}[\exp(J(x)/\beta)]$ is finite.
The KL-regularized tilting objective is
\begin{equation}
\label{eq:app_tilt}
\max_{q\,:\,q\ll p_{\cD}}\; \E_{x\sim q}[J(x)]-\beta\,\KL\bigl(q\|p_{\cD}\bigr).
\end{equation}
To see that the optimizer has the form $q^\star(x)\propto p_{\cD}(x)\exp(J(x)/\beta)$, write the Lagrangian with multiplier $\lambda$ for the normalization constraint:
\[
\mathcal{L}(q,\lambda)
=
\int q(x)J(x)\,dx
-\beta\int q(x)\log\frac{q(x)}{p_{\cD}(x)}\,dx
+\lambda\Bigl(1-\int q(x)\,dx\Bigr).
\]
Taking the functional derivative with respect to $q(x)$ and setting it to zero:
\[
J(x)-\beta\Bigl(1+\log\frac{q(x)}{p_{\cD}(x)}\Bigr)-\lambda=0.
\]
Rearranging gives $\log q(x)=\log p_{\cD}(x)+\frac{J(x)}{\beta}+c$, where $c$ collects constants.
Exponentiating yields $q^\star(x)\propto p_{\cD}(x)\exp(J(x)/\beta)$.
Instantiating $J(x_t)=T_t$ for PTR recovers the weight $w_t$ from Eq~\eqref{eq:weight}.

The temperature $\beta$ controls the trade-off between score maximization and proximity to the original data distribution.
As $\beta\to\infty$, the KL penalty dominates and $q^\star\to p_{\cD}$ (uniform weighting, equivalent to standard SFT).
As $\beta\to 0^+$, the tilted distribution concentrates on the highest-scoring samples.
PTR uses this exponential form as the raw score-to-weight map and then applies clipping and mixture as in Eq~\eqref{eq:weight}.
Within the unclipped regime this preserves the ordering induced by $T_t$; outside it, extreme ratios are intentionally flattened to enforce conservatism.
PTR operates in an intermediate regime where $\beta$ is adapted online by the adaptive scale controller (Section~\ref{sec:adarefiner}) to maintain a meaningful but bounded weight range.

\textbf{Selective transfer from cross-embodiment data.}
The exponential tilting derived above also induces a principled reallocation of effective weight across data sources.
Cross-embodiment and multi-operator datasets are naturally modeled as mixtures of source-specific distributions.
PTR provides both a floor and a ceiling for cross-embodiment learning.
The floor suppresses mismatched or suboptimal pooled data: when a helper embodiment's trajectories produce post-action observations that are uninformative for the target embodiment, the posterior stays diffuse and the sample loses extra emphasis, reverting toward uniform weighting and sometimes below it after exponentiation and clipping.
The ceiling amplifies useful transfer: when helper data cover task-relevant regions that the target embodiment lacks, the posterior sharpens and PTR allocates more credit there.
The following result, proved in Appendix~\ref{app:proofs}, formalizes this selective pooling at the source level.

\begin{cthproposition}{Source reweighting under exponential tilting}
\label{prop:mixture_main}
Let $p_{\cD}(x)=\sum_m \pi_m p_m(x)$ be a mixture over $M$ sources.
Assume also that $\E_{p_m}[\exp(J(x)/\beta)]<\infty$ for every source $m$.
Tilting by $\exp\!\bigl(J(x)/\beta\bigr)$ yields
\begin{equation}
\label{eq:mixture_main}
q^\star(m)
  =\frac{\pi_m\,\E_{p_m}\!\bigl[\exp(J(x)/\beta)\bigr]}
        {\sum_{j} \pi_j\,\E_{p_j}\!\bigl[\exp(J(x)/\beta)\bigr]}.
\end{equation}
\end{cthproposition}

Sources whose samples consistently receive high PTR scores gain effective weight in the induced training distribution; sources with low scores are suppressed.
The unified action space of Being-H0.5~\citep{luo2026being} ensures that cross-embodiment actions share a common representation, making the identification signal meaningful across embodiment boundaries.

\subsection{Conservative reweighting and the induced training distribution}
\label{sec:ptr_reweight}

\textbf{Conservative weight mapping.}
PTR maps the score to a conservative per-sample weight via exponentiation, clipping, and a convex mixture:
\begin{equation}
\label{eq:weight}
w_t := 1+\alpha\Bigl(\mathrm{clip}_{[w_{\min},w_{\max}]}\bigl(\exp(T_t/\beta)\bigr)-1\Bigr),
\qquad \alpha\in[0,1].
\end{equation}
The exponential form is the optimizer of the KL-regularized tilting objective derived in Section~\ref{sec:theory}.
Clipping and mixture control variance and bound distribution shift, similar in spirit to truncated importance weighting~\citep{schulman2017proximal,espeholt2018impala}.

\textbf{Density ratio bound.}
Define the implicit distribution $q(x)\propto p_{\cD}(x)\,w(x)$.
If $w(x)\in[\underline w, \overline w]$ for all $x$, then the density ratio is bounded:
\begin{equation}
\label{eq:ratio_bound}
\frac{\underline w}{\E_{p_{\cD}}[w]}
\ \le\
\frac{q(x)}{p_{\cD}(x)}
\ \le\
\frac{\overline w}{\E_{p_{\cD}}[w]}.
\end{equation}
Under the mixture form in Eq~\eqref{eq:weight} with $\alpha\in[0,1]$ and clipping $w\in[w_{\min},w_{\max}]$, the effective bounds are $\underline w=1+\alpha(w_{\min}-1)$ and $\overline w=1+\alpha(w_{\max}-1)$.
Setting $r_{\max}=\overline w / \underline w$, the KL divergence satisfies $\KL(q\|p_{\cD})\le \log r_{\max}$.
With the default PTR parameters ($w_{\min}=0.25$, $w_{\max}=4.0$, $\alpha=1$), this gives $\KL(q\|p_{\cD})\le \log 16\approx 2.77$ nats.
This bound is the formal justification for calling PTR ``conservative'': the induced training distribution can never drift far from the original data distribution, regardless of the score function.

\textbf{Clipping and mixture bound: detailed derivation.}
Define the implicit distribution $q(x)\propto p_{\cD}(x) w(x)$ from Eq~\eqref{eq:implicit_q}, where $w(x)$ is the clipped-and-mixed weight.
If $w(x)\in[\underline w, \overline w]$ for all $x$, then the density ratio is bounded:
\begin{equation}
\label{eq:app_ratio_bound}
\frac{\underline w}{\E_{p_{\cD}}[w]}
\ \le\
\frac{q(x)}{p_{\cD}(x)}
\ \le\
\frac{\overline w}{\E_{p_{\cD}}[w]}.
\end{equation}
Under the mixture form in Eq~\eqref{eq:weight} with $\alpha\in[0,1]$ and clipping $w\in[w_{\min},w_{\max}]$, the effective bounds are
\begin{equation}
\label{eq:app_eff_bounds}
\underline w=1+\alpha(w_{\min}-1),\qquad
\overline w=1+\alpha(w_{\max}-1).
\end{equation}
In the minibatch implementation, $\E_{p_{\cD}}[w]$ is replaced by the empirical denominator $\sum_t w_t$ from Eq~\eqref{eq:weighted_loss}; the same pointwise bounds hold once all per-sample weights lie in the prescribed interval.

By definition, $q(x)=p_{\cD}(x) w(x)/Z$ with normalizer $Z=\E_{p_{\cD}}[w]$.
Dividing both sides by $p_{\cD}(x)$:
\begin{equation}
\frac{q(x)}{p_{\cD}(x)}
=
\frac{w(x)}{Z}.
\end{equation}
Since $w(x)\in[\underline w, \overline w]$ for all $x$, and $Z$ is a fixed positive constant, the ratio is bounded by $\underline w / Z$ from below and $\overline w / Z$ from above, yielding Eq~\eqref{eq:app_ratio_bound}.

\textbf{KL bound.}
Setting $r_{\max}=\overline w / \underline w$, the pointwise bound implies
\begin{equation}
\KL(q\|p_{\cD})
=
\E_q\!\left[\log\frac{q(x)}{p_{\cD}(x)}\right]
\le
\log r_{\max}
=
\log\frac{\overline w}{\underline w}.
\end{equation}
With the default PTR parameters ($w_{\min}=0.25$, $w_{\max}=4.0$, $\alpha=1$), we obtain $\KL(q\|p_{\cD})\le \log 16\approx 2.77$ nats.
When $\alpha=1$, the mixture form reduces to pure clipping, so $\underline w = w_{\min}$ and $\overline w = w_{\max}$.
Smaller $\alpha$ tightens the bound further toward zero.
The general $\alpha\in[0,1]$ formulation in Eq~\eqref{eq:app_eff_bounds} is retained because it clarifies the continuum between uniform SFT ($\alpha=0$) and full PTR reweighting ($\alpha=1$), and because practitioners working with smaller or noisier datasets may benefit from intermediate values.

The clipping range $[w_{\min}, w_{\max}]$ and mixture coefficient $\alpha$ are the two user-facing knobs that control the conservatism level.
In the limit $\alpha\to 0$, PTR recovers uniform SFT with $q=p_{\cD}$.

\textbf{Self-normalized weighted regression.}
PTR applies the weight to the original supervised action loss:
\begin{equation}
\label{eq:weighted_loss}
\mathcal{L}^{\star}_{\mathrm{act}}
=
\frac{\sum_{t}\sg{w_t}\,\ell_{\mathrm{act}}(\phi;h_t,s_t,a_{t:t+L-1})}{\sum_t \sg{w_t}}.
\end{equation}
The stop-gradient on $w_t$ blocks the path where the policy could increase its own weights by manipulating the scorer.
The scorer still learns from its own identification loss. Eq~\eqref{eq:weighted_loss} also gives a direct distribution view.
Let $x_t$ denote the full chunk-level sample excluding the post-action observation, \eg $x_t=(o_t,s_t,l,a_{t:t+L-1})$.
Self-normalized weighting makes the gradient update equivalent to sampling from an implicit distribution
\begin{equation}
\label{eq:implicit_q}
q(x)\ \propto\ p_{\cD}(x)\,w(x).
\end{equation}
Clipping and mixture bound how far $q$ can deviate from $p_{\cD}$ (Eq~\eqref{eq:ratio_bound}).
This is the operational meaning of \emph{conservative} in PTR.

\textbf{Mixture lens for selective pooling.}
Proposition~\ref{prop:mixture_main} formalizes how tilting reallocates weight across sources: setting $J(x_t)=T_t$ (the PTR score), the ratio $q^\star(m)/\pi_m$ is proportional to the moment-generating function $\E_{p_m}[\exp(T_t/\beta)]$ of the score distribution within source $m$.
Sources whose samples consistently yield high PTR scores have large moment-generating values and gain effective weight.
Sources dominated by ambiguous samples remain closer to their original proportion, while sources with persistently low or negative scores are suppressed.
This provides a formal mechanism for the ``floor and ceiling'' described above: the floor suppresses low-quality or mismatched sources, while the ceiling amplifies sources that provide informative coverage for the target embodiment.

\textbf{Conservative bounds at the source level.}
In practice, PTR applies clipping and mixture (Eq~\eqref{eq:weight}), so the effective per-sample weight satisfies $w(x)\in[\underline w,\overline w]$.
The induced source proportion becomes
\begin{equation}
    q(m)=\frac{\pi_m}{Z}\E_{x\sim p_m}[w(x)],
\end{equation}
where $Z=\E_{p_{\cD}}[w]$.
Because every per-sample weight lies in $[\underline w,\overline w]$, the ratio $q(m)/\pi_m = \E_{p_m}[w]/Z$ is bounded:
\begin{equation}
\label{eq:source_bound}
\frac{\underline w}{\overline w}
\ \le\
\frac{q(m)}{\pi_m}
\ \le\
\frac{\overline w}{\underline w}.
\end{equation}
With the default PTR parameters ($w_{\min}=0.25$, $w_{\max}=4.0$, $\alpha=1$), this gives $q(m)/\pi_m \in [1/16, 16]$, and the mixture coefficient $\alpha<1$ tightens this range further toward unity.
No source can be amplified or suppressed by more than a bounded factor, regardless of how extreme its score distribution is.

\textbf{Implementation considerations.}
Real datasets can record post-action observations at different offsets $\Delta$.
Large offsets can make identification harder and inject additional ambiguity.
To remain conservative across offsets, we apply a discount $\gamma^{\Delta'}$ to the identification loss and, optionally, to the resulting weights, where $\Delta'$ is a normalized offset and $\gamma\in(0,1]$.
In large heterogeneous robot corpora, some mismatched targets can still be semantically close to the positive, especially under cross-embodiment pooling or repeated task motifs.
PTR treats these cases conservatively: overlap lowers the identification margin and pushes the posterior toward uniform rather than creating spuriously large weights.
Together with the fallback $w_t=1$ for missing targets, uncertainty primarily removes extra emphasis instead of fabricating it.

\subsection{Adaptive scale control}
\label{sec:adarefiner}

The identification posterior is informative only when it is neither saturated nor completely flat.
The adaptive scale controller is a multi-system training feedback mechanism~\citep{zhang2024adarefiner} that monitors identification statistics at each logging step and adjusts three parameters within fixed bounds to keep training in a stable operating range.

\textbf{Monitored statistics.}
The refiner tracks exponential moving averages (EMA, decay $0.98$) of four quantities:
(i)~identification accuracy (\texttt{nce\_acc}), the fraction of samples where the matched target receives the highest logit;
(ii)~score margin (\texttt{nce\_margin}), the gap between the matched logit and the best mismatched logit;
(iii)~mean PTR score (\texttt{nce\_adv}); and
(iv)~the valid-target ratio, the fraction of samples with a usable post-action observation.

\textbf{Adapted parameters.}
Based on these statistics, the refiner adjusts:
\begin{itemize}
\item \emph{Scorer temperature} $\tau_{\mathrm{score}}$: raised when \texttt{nce\_acc} falls below a lower threshold (to keep an immature scorer conservative), and lowered when it exceeds an upper threshold (to sharpen a mature scorer). Bounded in $[\tau_{\min}, \tau_{\max}]=[0.03,\,0.20]$.
\item \emph{Advantage scaling} $\beta$: adjusted to keep the effective weight range meaningful. Larger $\beta$ makes weights more conservative by compressing them toward one, while smaller $\beta$ allows stronger differentiation once the scorer is reliable. Bounded in $[\beta_{\min}, \beta_{\max}]=[0.5,\,3.0]$.
\item \emph{Hard-negative ratio}: the fraction of candidate-pool negatives drawn from semantically similar samples (same task family or similar proprioceptive state) rather than uniformly at random. Increased as the scorer matures (rising \texttt{nce\_acc}) to maintain a challenging identification task. Bounded in $[0,\,0.5]$.
\end{itemize}

\textbf{Update rule.}
Let $\bar T$ denote the EMA-smoothed identification accuracy.
The refiner applies two conditional branches at each logging step.
If $\bar T < 0.05$ (scorer too weak), both $\tau_{\mathrm{score}}$ and $\beta$ are multiplied by $1.01$ to keep the posterior and weights conservative while the scorer is still organizing the candidate space.
If $\bar T > 0.35$ and the score margin exceeds $0.10$ (scorer confident and well-separated), both parameters are multiplied by $0.995$ to sharpen the posterior and allow stronger differentiation between informative and ambiguous samples.
The hard-negative ratio is set by linear interpolation of $\bar T$ within $[0.10,\,0.50]$: when $\bar T \le 0.10$ the ratio is $0.0$ (all random negatives), and when $\bar T \ge 0.50$ the ratio saturates at $0.5$.
All updates are clipped to the parameter's allowed range after each step.
These adjustments change only scalar sharpness and sampling parameters; for a fixed candidate set they can widen or narrow the posterior and the induced weight spread, but they do not alter which target is designated as matched.

\textbf{Empirical behavior.}
Training dynamics of the belief tokenizer and adaptive scale controller are visualized in Figures~\ref{fig:training_analysis}--\ref{fig:hyperparam}.
Two observations from the controller logs are worth noting.
First, $\tau_{\mathrm{score}}$ decreases from its initial value to approximately $0.03$ as the contrastive signal strengthens, sharpening the posterior.
Second, the hard-negative ratio increases from $0.0$ to $0.50$ over training, indicating that the refiner progressively enriches the candidate pool with challenging negatives as the scorer matures.

\subsection{Practical pipeline and gradient routing}
\label{sec:ptr_pipeline}

\textbf{Modules.}
PTR augments a standard post-training stack with four lightweight components:
an EMA target encoder $g$ for post-action targets; a query head $f$; a candidate pool built from in-batch targets, cross-rank targets, and a FIFO queue; and a BeliefTokenizer $B$ that produces belief proxy tokens.

\textbf{Estimator losses.}
PTR trains the scorer with the identification loss
\begin{equation}
\label{eq:nce}
\mathcal L_{\mathrm{id}}(\theta)
=
-\E_t\Bigl[\log \hat p(I_t=0\mid h_t,e_t,Y_t)\Bigr],
\end{equation}
and an action-sensitivity regularizer (the ranking loss $\mathcal L_{\mathrm{rank}}$ defined in Section~\ref{sec:ptr_score}).
The BeliefTokenizer uses entropy and diversity regularizers on its soft assignments (Section~\ref{sec:belief_tokenizer}).

\textbf{One-step optimization with routed gradients.}
The total objective is
\begin{equation}
\label{eq:total_loss}
\mathcal L_{\mathrm{total}}
=
\mathcal{L}^{\star}_{\mathrm{act}}
+\lambda_{\mathrm{id}}\mathcal L_{\mathrm{id}}
+\lambda_{\mathrm{rank}}\mathcal L_{\mathrm{rank}}
+\mathcal L_{\mathrm{tok}}.
\end{equation}
All trainable components are updated jointly.
Stop-gradient markers block specific paths (Figure~\ref{fig:framework}):
(i) $\sg{w_t}$ blocks the policy$\rightarrow$scorer manipulation path, but the scorer still learns from $\mathcal L_{\mathrm{id}}$ and $\mathcal L_{\mathrm{rank}}$;
(ii) $\sg{g(\cdot)}$ makes the post-action target encoder an EMA teacher rather than an online head;
(iii) $\sg{z_{t+1}}$ blocks gradients through time, while $B$ still learns from current-step losses;
(iv) the candidate queue and cross-rank gather are non-differentiable, but the scorer still receives gradients on the current query and the current matched target.
Conservative weights $w_t$ are applied only to the action loss; the remaining terms receive uniform gradients.

Since $w_t$ is treated as a constant, PTR does not alter attention computations.
It only rescales per-sample gradients.
Let $\bar w_t=\sg{w_t}/\sum_j\sg{w_j}$.
Then for any shared parameter block $\phi$ in the backbone or action head,
\begin{equation}
\label{eq:grad_scale}
\nabla_\phi\,\mathcal{L}^{\star}_{\mathrm{act}}=\sum_t \bar w_t\,\nabla_\phi \ell_{\mathrm{act}}(\phi;h_t,s_t,a_{t:t+L-1}).
\end{equation}

\textbf{Transformer view.}
PTR assumes the backbone is a transformer that consumes a token sequence under a fixed attention mask (causal on language tokens, bidirectional within vision tokens).
Belief proxy tokens are inserted as ordinary context tokens and follow the same mask semantics.
PTR does not modify the mask; its only interaction with the transformer update is the gradient scaling in Eq~\eqref{eq:grad_scale}.

\textbf{Algorithmic summary.}
Algorithm~\ref{alg:ptr} summarizes one PTR training segment.
The outer loop iterates over chunks within a trajectory segment; the inner computation at each chunk produces a conservative weight $w_t$ that rescales the action loss gradient.
All modules (scorer, BeliefTokenizer, action head) share a single forward/backward pass, so the algorithm adds no extra environment interaction or second-stage optimization.

\begin{algorithm}[ht]
\caption{Posterior-Transition Reweighting (PTR) for one segment}
\label{alg:ptr}
\begin{algorithmic}[1]
\State Initialize belief tokens $z_{t_0}\gets z_{\mathrm{init}}$
\For{$t=t_0,\ldots,t_0+T-1$}
    \State Compute $(H_t,h_t)=f_\phi(o_t,l,z_t)$
    \State Compute action loss $\ell_{\mathrm{act}}(\phi;h_t,s_t,a_{t:t+L-1})$ and action tokens $E_t=g_\phi(h_t,s_t,a_{t:t+L-1})$; set $e_t=\mathrm{pool}(E_t)$
    \State Compute post-action target $y_{t,0}^{+}=\sg{g(o_{t+\Delta})}$
    \State Build candidate set $Y_t=\{y_{t,0}^{+}\}\cup Y_t^{-}$ from in-batch / cross-rank / queue
    \State Compute query $u_t=f(h_t,e_t)$ and posterior $\hat p(I_t=0\mid h_t,e_t,Y_t)$
    \State Compute score $T_t=\log \frac{\hat p(I_t=0\mid h_t,e_t,Y_t)}{1/|Y_t|}$
    \State Compute conservative weight $w_t=1+\alpha(\mathrm{clip}(e^{T_t/\beta})-1)$
    \State Update belief tokens $z_{t+1}=\sg{B(H_t,E_t)}$
\EndFor
\State Update parameters by minimizing $\mathcal L_{\mathrm{total}}$ in Eq~\eqref{eq:total_loss} with $\sg{w_t}$
\end{algorithmic}
\end{algorithm}

\section{Experiments}
\label{sec:exp}

\subsection{Setup and baselines}
\label{sec:exp_setup}

All methods build on Being-H0.5~\citep{luo2026being}, a state-of-the-art VLA whose $200$-dimensional unified action space maps heterogeneous robots to shared semantic slots, making it a natural testbed for cross-embodiment post-training.
All methods share the same backbone, action head, and training budget ($60$k steps, batch size $128$, cosine learning-rate schedule; full hyperparameter listing in Appendix~\ref{app:hyperparams}).
Three configurations are compared throughout:
(i)~SFT sets $\alpha{=}0$ in \cref{eq:weight}, recovering uniform supervised post-training;
(ii)~SFT+Belief adds belief proxy tokens but keeps uniform weighting, isolating the effect of richer context from the effect of reweighting;
(iii)~PTR enables the full pipeline with $\alpha{=}1$.

Two simulation benchmarks complement the real-robot evaluation.
LIBERO~\citep{liu2023libero} provides four task suites (Spatial, Object, Goal, Long-Horizon), each containing $10$ tasks evaluated over $50$ episodes per task ($500$ per suite, $2000$ total).
RoboCasa~\citep{nasiriany2024robocasa} provides $24$ kitchen tasks evaluated over $50$ trials across $5$ scene layouts per task ($1200$ total).

To stress-test robustness, we introduce four training-data corruptions applied before post-training:
(i)~Action Noise Injection (ANI) adds Gaussian noise $\epsilon\sim\mathcal{N}(0,\sigma^2 I)$ with $\sigma{=}0.1$ to $30$\% of trajectories, simulating teleoperation jitter;
(ii)~Trajectory Truncation (TT) randomly truncates $25$\% of trajectories to $40$--$70$\% of their original length, simulating incomplete demonstrations;
(iii)~Label Noise (LN) randomly reassigns language instructions for $20$\% of trajectories, simulating annotation errors;
(iv)~Combined applies all three simultaneously.
All corruptions are applied only to the post-training data; the evaluation environments and their success criteria are left unchanged.

\subsection{Simulation benchmarks}
\label{sec:exp_sim}

\begin{table}[t]
\centering
\caption{Standard simulation results (success rate \%) on LIBERO and RoboCasa. LIBERO reports per-suite averages over $500$ episodes each; RoboCasa reports category averages over $1200$ total trials. \textbf{Bold}: best; \underline{Underlined}: second best.}
\label{tab:standard}
\setlength{\tabcolsep}{4pt}
\begin{tabular}{l cccc c ccc c}
\toprule
& \multicolumn{4}{c}{\textbf{LIBERO}} & & \multicolumn{3}{c}{\textbf{RoboCasa}} & \\
\cmidrule(lr){2-5}\cmidrule(lr){7-9}
\textbf{Method} & Spatial & Object & Goal & Long & \textbf{Avg} & Pick\&Pl. & Door/Dr. & Others & \textbf{Avg} \\
\midrule
SFT & \textbf{98.8} & \underline{99.0} & \textbf{98.6} & \underline{96.8} & \textbf{98.3} & 36.0 & 71.3 & \underline{55.3} & 54.2 \\
SFT+Belief & \underline{98.2} & 98.4 & \underline{98.0} & 95.6 & 97.6 & \underline{36.7} & \underline{71.5} & 55.0 & \underline{54.4} \\
PTR & 98.0 & \textbf{99.2} & 97.6 & \textbf{97.0} & \underline{97.8} & \textbf{38.3} & \textbf{73.0} & \textbf{55.5} & \textbf{55.6} \\
\bottomrule
\end{tabular}
\end{table}

\begin{figure}[ht]
    \centering
    \includegraphics[width=\linewidth]{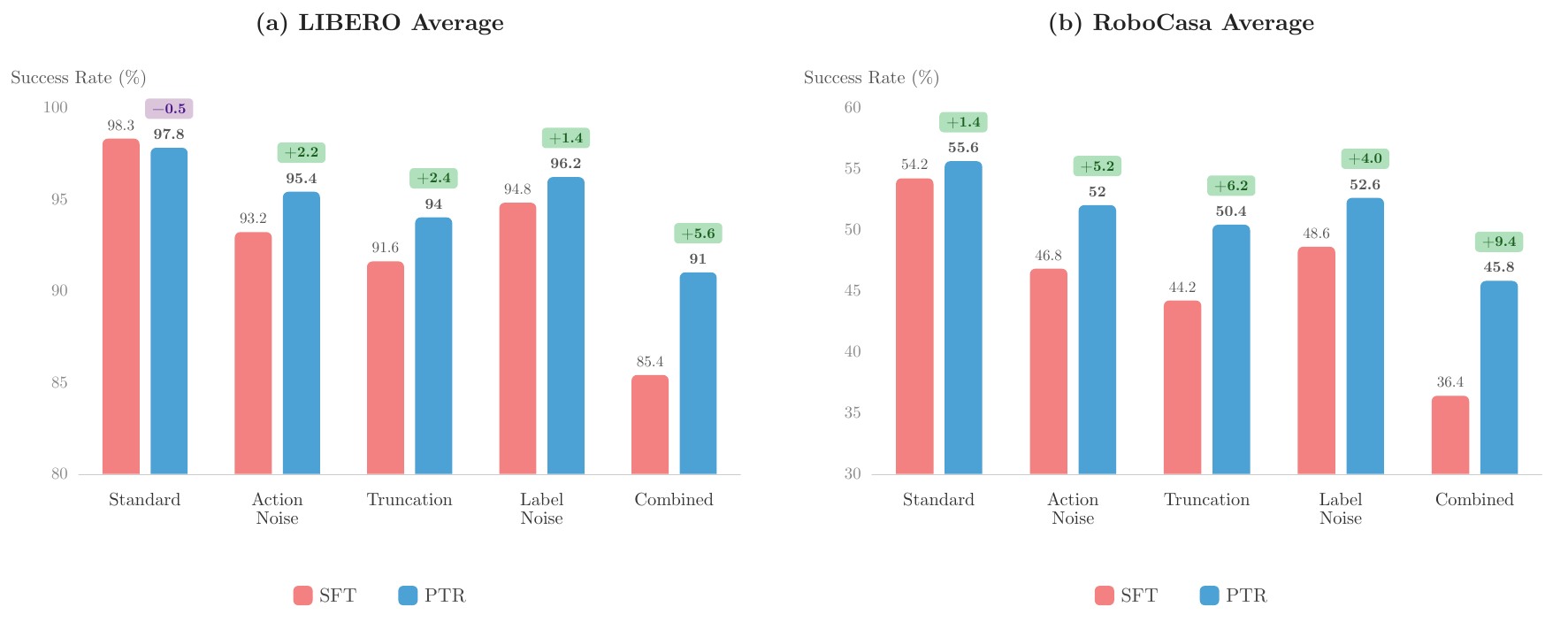}
    \caption{Robustness under corrupted training data (success rate \%). Colored badges show PTR$-$SFT deltas. PTR maintains higher performance across all corruption types.}
    \label{fig:corrupted_bar}
\end{figure}

\textbf{Standard evaluation.}
Table~\ref{tab:standard} reports results on curated (uncorrupted) training data.
On LIBERO, all three methods remain near ceiling ($97.6$--$98.3$\%), indicating that PTR stays competitive when the training data are already clean.
PTR leads on Object ($99.2$) and Long-Horizon ($97.0$), the two suites where additional selectivity appears most helpful.

On RoboCasa, where demonstrations span diverse kitchen scenes and object configurations, PTR outperforms SFT by $1.4$ pp ($55.6$ vs.\ $54.2$), with gains across all three task categories: Pick\&Place ($38.3$ vs.\ $36.0$), Door/Drawer ($73.0$ vs.\ $71.3$), and Others ($55.5$ vs.\ $55.3$).

SFT+Belief falls between the two on RoboCasa ($54.4$), indicating that belief tokens improve context quality but the reweighting signal adds further value on heterogeneous benchmarks.
The gap between SFT+Belief and PTR ($1.2$ pp) isolates the contribution of the identification-based reweighting beyond what richer context alone provides.

\subsection{Robustness under corrupted training data}
\label{sec:exp_corrupt}

Figure~\ref{fig:corrupted_bar} compares PTR and SFT under the four corruption protocols.
Under Action Noise Injection, SFT drops by $5.1$ pp on LIBERO while PTR drops by only $2.4$ pp, because noisy action chunks produce atypical post-action consequences that the identification posterior tends to de-emphasize.

Trajectory Truncation and Label Noise show a similar pattern. On LIBERO, SFT loses $6.7$ pp and $3.5$ pp, while PTR loses $3.8$ pp and $1.6$ pp. On RoboCasa, the corresponding drops are $10.0$ pp and $5.6$ pp for SFT versus $5.2$ pp and $3.0$ pp for PTR.
The combined corruption is the most revealing: SFT loses $12.9$ pp on LIBERO and $17.8$ pp on RoboCasa relative to the clean baseline, whereas PTR loses $6.8$ pp and $9.8$ pp, yielding absolute gains of \emph{$+5.6$ pp} and \emph{$+9.4$ pp} respectively.

The pattern is consistent across both benchmarks: corrupted samples typically produce less concentrated identification posteriors and therefore lose relative emphasis, while clean samples with distinctive consequences are amplified.
RoboCasa shows larger absolute gaps than LIBERO under all corruption types, reflecting its greater inherent heterogeneity in kitchen scenes and object configurations.

\begin{figure}[t]
\centering
    \includegraphics[width=\linewidth]{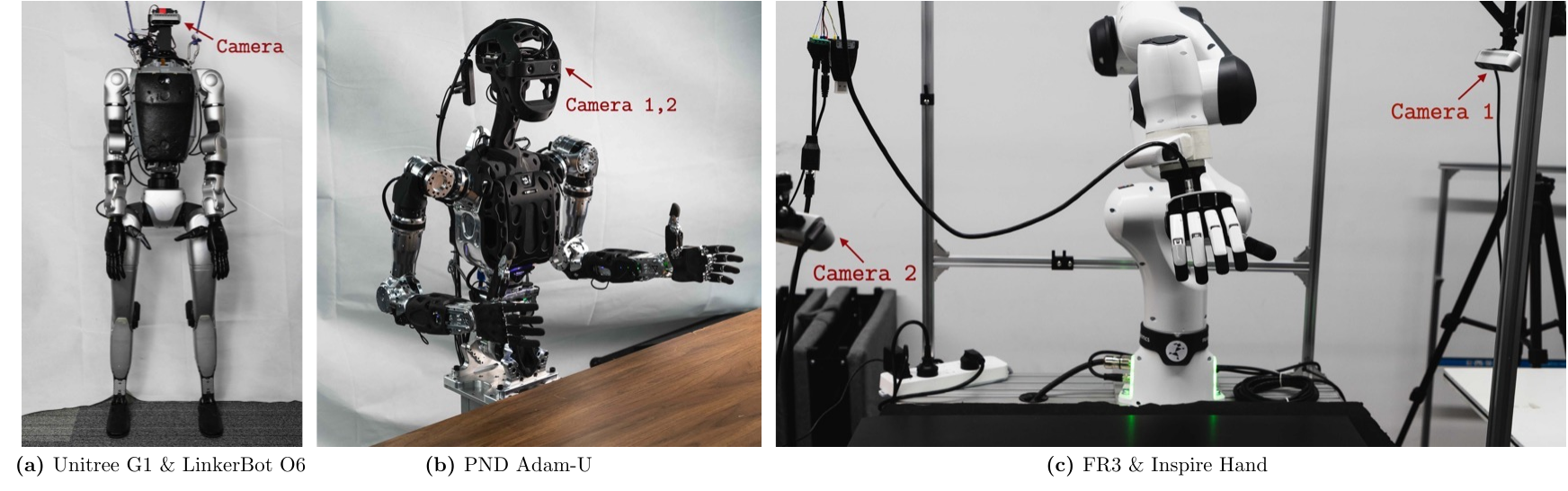}
    \caption{Real-robot platforms used for evaluation.
    (a)~Unitree G1 with LinkerHand O6 dexterous hands.
    (b)~PND Adam-U with bimanual dexterous manipulation and a movable head.
    (c)~FR3 single-arm with Inspire dexterous hand.}
    \label{fig:real-robots}
\end{figure}

\begin{table}[ht]
\centering
\caption{Robot platform specifications for the three real-robot embodiments.}
\label{tab:robot_specs}
\setlength{\tabcolsep}{5pt}
\resizebox{\linewidth}{!}{
\begin{tabular}{l cccccc}
\toprule
\textbf{Platform} & \textbf{DoF} & \textbf{DoF breakdown} & \textbf{Morphology} & \textbf{Camera} & \textbf{View} & \textbf{Hz} \\
\midrule
Unitree G1 + LinkerHand O6 & $26$ & $7{\times}2$ arms $+$ $6{\times}2$ hands & Bimanual, dexterous & D435 & Fixed ego & $20$ \\
PND Adam-U & $31$ & $3$ waist $+$ $2$ head $+$ $7{\times}2$ arms $+$ $6{\times}2$ hands & Bimanual + head + waist & ZED Mini & Movable ego (dual) & $20$ \\
FR3 + Inspire Hand & $13$ & $7$ arm $+$ $6$ hand & Single-arm, dexterous & $2{\times}$D435 & Fixed 3rd-person & $50$ \\
\bottomrule
\end{tabular}
}
\end{table}

\subsection{Real-robot evaluation and cross-embodiment transfer}
\label{sec:exp_real}

Three real-robot platforms span a range of embodiment complexity.
Table~\ref{tab:robot_specs} summarizes their specifications.

\textbf{Unitree G1 + LinkerHand O6.}
A $26$-DoF bimanual humanoid upper-body setup: $7$ joints per arm ($\times 2$) and $6$ joints per dexterous hand ($\times 2$), with a single fixed egocentric Intel RealSense D435 camera running at $20$\,Hz.
Its tasks emphasize dual-hand coordination, regrasping, and larger workspace transfers.

\textbf{PND Adam-U.}
A $31$-DoF bimanual platform: $3$ waist joints, $2$ head joints, $7$ joints per arm ($\times 2$), and $6$ joints per dexterous hand ($\times 2$), with dual ego views from a movable head-mounted ZED Mini stereo rig running at $20$\,Hz.
Compared with G1, it introduces stronger viewpoint drift due to head motion and longer multistage sequences enabled by the additional waist and head.

\textbf{FR3 + Inspire Hand.}
A $13$-DoF single-arm platform: $7$ arm joints and $6$ hand joints, with two fixed third-person Intel RealSense D435 cameras running at $50$\,Hz.
It stresses precise single-arm grasping, placement, and contact-sensitive manipulation under tighter kinematic constraints.

These three embodiments create the heterogeneity that PTR is designed to handle: identical semantic progress can arise from very different camera trajectories, contact patterns, and kinematic solutions.

\textbf{Unified action space slot assignment.}
All three platforms share the $200$-dimensional unified action space of Being-H0.5~\citep{luo2026being}, but each embodiment activates only the semantic slots corresponding to its available motor components.
All platforms use the shared right-arm and right-hand semantic groups.
G1 additionally activates the mirrored left-arm and left-hand groups, yielding $26$ active dimensions in total.
Adam-U further activates head and waist groups on top of the bimanual groups, yielding $31$ active dimensions.
FR3 uses only the single-arm plus dexterous-hand groups, yielding $13$ active dimensions.
All remaining slots are masked to zero during both training and inference, so the flow-matching head operates only in the embodiment-relevant subspace while the shared semantic layout still supports cross-embodiment transfer.

\begin{figure}[t]
    \centering
    \includegraphics[width=\linewidth]{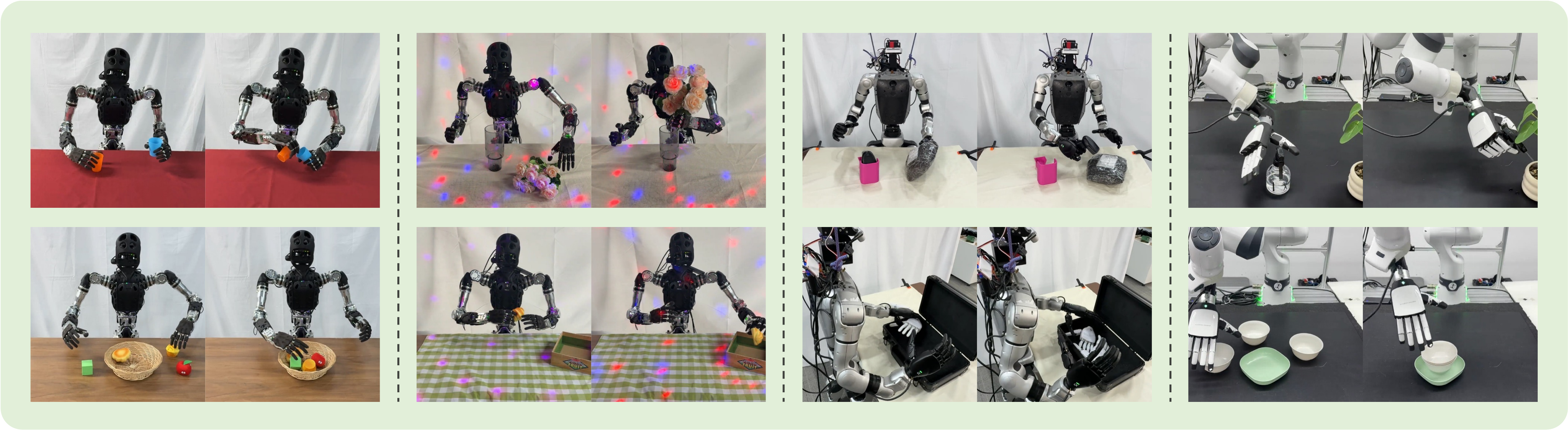}
    \caption{Representative real-robot tasks across three platforms and four capability suites.}
    \label{fig:real_tasks}
\end{figure}

\textbf{Evaluation protocol.}
Table~\ref{tab:cross_embodiment} consolidates all results across simulation and real-robot benchmarks under both specialist (per-embodiment) and generalist (single-checkpoint) training.
Each real-robot task is evaluated over $20$ trials.
$12$ tasks across three platforms are grouped into four capability-oriented suites, each containing three tasks ($60$ trials per suite, $240$ total):
Bimanual (dual-arm coordination), Long-Horizon (multi-step sequential manipulation), Spatial (precise placement and arrangement), and Robust (generalization under scene variation and contact uncertainty).
Figure~\ref{fig:real_tasks} illustrates representative tasks from each suite. Suite labels follow the evaluation protocol rather than the task title alone, so semantically related tasks can appear in different suites across embodiments when their dominant difficulty differs.

\begin{table}[t]
\centering
\caption{Specialist and generalist training results (success rate \%). Generalist trains a single checkpoint on all five sources (three real-robot sources plus LIBERO and RoboCasa); Specialist trains one checkpoint per source. ``Overall'' is the mean of the LIBERO average, the RoboCasa average, and the real-robot overall average. \textbf{Bold}: best; \underline{Underlined}: second best per column.}
\label{tab:cross_embodiment}
\setlength{\tabcolsep}{3pt}
\begin{tabular}{l cc cccc c}
\toprule
& \multicolumn{2}{c}{\textbf{Simulation}} & \multicolumn{4}{c}{\textbf{Real Robot}} & \\
\cmidrule(lr){2-3}\cmidrule(lr){4-7}
\textbf{Method} & LIBERO & RoboCasa & Bimanual & Long-Hor. & Spatial & Robust & \textbf{Overall} \\
\midrule
SFT-Specialist & \textbf{98.3} & 54.2 & 55.0 & \underline{63.3} & \underline{75.0} & 50.0 & 71.1 \\
PTR-Specialist & \underline{97.8} & \textbf{55.6} & \textbf{66.7} & 61.7 & \textbf{78.3} & \textbf{61.7} & \textbf{73.5} \\
SFT-Generalist & 96.2 & 50.8 & 45.0 & 51.7 & 63.3 & 40.0 & 65.7 \\
PTR-Generalist & 97.4 & \underline{55.0} & \underline{60.0} & \textbf{65.0} & 73.3 & \underline{56.7} & \underline{72.1} \\
\bottomrule
\end{tabular}
\end{table}

\textbf{Specialist results.}
Under per-embodiment training, PTR outperforms SFT on three of four real-robot suites.
The largest suite-level gains appear on Bimanual and Robust, both at $+11.7$ pp ($66.7$ vs.\ $55.0$ and $61.7$ vs.\ $50.0$ respectively), where operator variability and partial completions leave more room for the identification posterior to discriminate between informative and ambiguous chunks.
Spatial improves by $+3.3$ pp ($78.3$ vs.\ $75.0$).
On Long-Horizon, SFT retains a slight edge ($63.3$ vs.\ $61.7$), likely because sequential tasks exhibit lower operator variability and the identification signal provides less additional discrimination.
Across all $12$ tasks, PTR averages $67.1$\% versus $60.8$\% for SFT ($+6.3$ pp).

\begin{figure}[t]
    \centering
    \includegraphics[width=\linewidth]{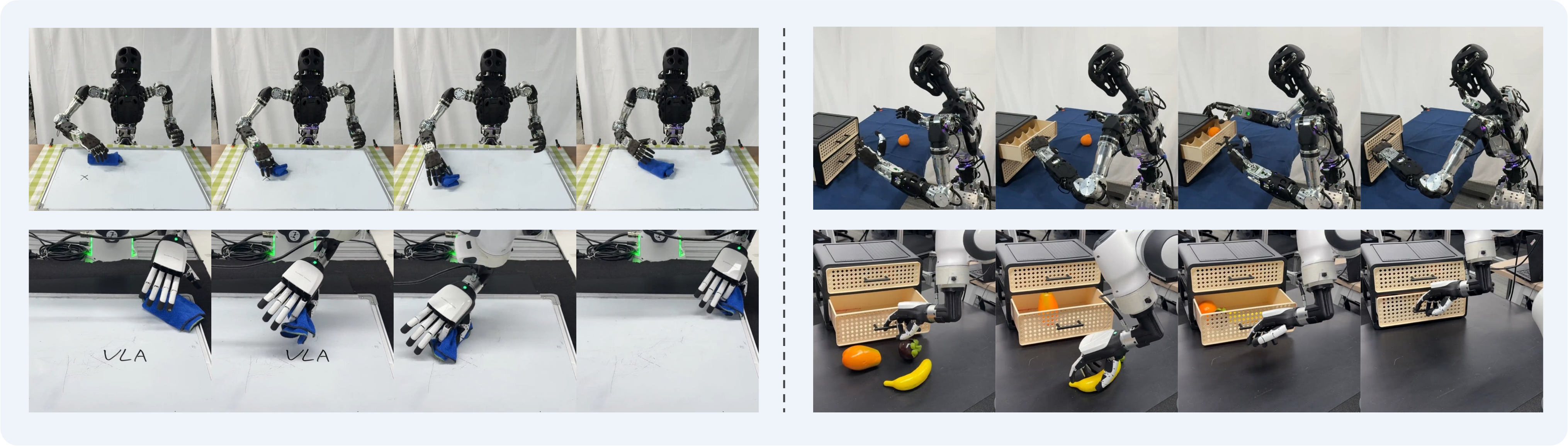}
    \caption{Cross-embodiment task correspondence. Different robot platforms (Adam-U \& FR3) execute semantically similar manipulation tasks, illustrating the shared post-action structure that enables PTR to selectively transfer useful knowledge.}
    \label{fig:cross_emb_tasks}
\end{figure}

\textbf{Cross-embodiment generalist training.}
Generalist training pools all five sources into a single checkpoint.
Under SFT, this degrades the real-robot average by $10.8$ pp due to embodiment conflicts.
PTR-Generalist limits the drop to $3.3$ pp, widening the PTR-vs-SFT gap from $+6.3$ pp (specialist) to $+13.8$ pp (generalist) on real robots.
Two results highlight selective transfer.
First, PTR-Generalist ($65.0$) \emph{surpasses} SFT-Specialist ($63.3$) on Long-Horizon, showing that cross-embodiment data provide useful coverage for multi-step tasks when filtered by PTR.
Second, on RoboCasa PTR-Generalist ($55.0$) approaches PTR-Specialist ($55.6$) with only a $0.6$ pp drop, versus $3.4$ pp for SFT.
On LIBERO, PTR-Generalist ($97.4$) nearly matches SFT-Specialist ($98.3$), confirming that PTR extracts useful signal from cross-embodiment data even when the specialist already saturates.
These results align with Proposition~\ref{prop:mixture_main}: PTR concentrates effective weight on samples whose post-action consequences overlap with the target embodiment, suppressing mismatched sources while amplifying informative ones.
Figure~\ref{fig:cross_emb_tasks} visualizes this overlap across platforms.

The specialist-versus-generalist comparison reveals two patterns.
Even without cross-embodiment pooling, the specialist comparison still favors PTR on average, confirming that identification-based reweighting provides value beyond cross-embodiment transfer by suppressing ambiguous or low-quality samples within a source's own data distribution.
The PTR--SFT improvement is consistently larger in the generalist setting, where Proposition~\ref{prop:mixture_main} provides explanatory power: helper samples from other sources receive high PTR scores only when their post-action consequences are identifiable under the target's context representation.

\textbf{Per-task breakdown.}
Table~\ref{tab:per_task_real} breaks down the real-robot results into individual tasks.
PTR improves over SFT on $8$ of $12$ tasks in the specialist setting, ties on $3$, and trails on $1$, confirming that the aggregate gain is broad rather than driven by a single outlier task.
In the generalist setting, PTR improves over SFT on all $12$ tasks.
The generalist PTR checkpoint exceeds the specialist SFT checkpoint on $7$ of $12$ tasks and matches it on $4$ more, demonstrating that selective cross-embodiment pooling can partially compensate for the dilution of multi-source training.

\begin{table}[ht]
\centering
\caption{Per-task real-robot success rates (\%).}
\label{tab:per_task_real}
\small
\setlength{\tabcolsep}{3.5pt}
\resizebox{.8\linewidth}{!}{
\begin{tabular}{l l cccc}
\toprule
\textbf{Suite / Task} & \textbf{Platform} & \textbf{SFT-Spec} & \textbf{PTR-Spec} & \textbf{SFT-Gen} & \textbf{PTR-Gen} \\
\midrule
\multicolumn{6}{l}{\emph{Bimanual}} \\
Pour Water & Adam-U & $60.0$ & $70.0$ & $50.0$ & $65.0$ \\
Handover & Adam-U & $50.0$ & $65.0$ & $40.0$ & $55.0$ \\
Put \& Close Box & G1 & $55.0$ & $65.0$ & $45.0$ & $60.0$ \\
\rowcolor{gray!12} \multicolumn{2}{l}{\emph{Suite average}} & $55.0$ & $66.7$ & $45.0$ & $60.0$ \\
\midrule
\multicolumn{6}{l}{\emph{Long-Horizon}} \\
Clear Desk & Adam-U & $70.0$ & $65.0$ & $60.0$ & $70.0$ \\
Drawer Organization & Adam-U & $60.0$ & $60.0$ & $45.0$ & $65.0$ \\
Scan Package & G1 & $60.0$ & $60.0$ & $50.0$ & $60.0$ \\
\rowcolor{gray!12} \multicolumn{2}{l}{\emph{Suite average}} & $63.3$ & $61.7$ & $51.7$ & $65.0$ \\
\midrule
\multicolumn{6}{l}{\emph{Spatial}} \\
Arrange Flower & Adam-U & $80.0$ & $85.0$ & $70.0$ & $80.0$ \\
Stack Bowl & FR3 & $70.0$ & $75.0$ & $60.0$ & $70.0$ \\
Water Plant & FR3 & $75.0$ & $75.0$ & $60.0$ & $70.0$ \\
\rowcolor{gray!12} \multicolumn{2}{l}{\emph{Suite average}} & $75.0$ & $78.3$ & $63.3$ & $73.3$ \\
\midrule
\multicolumn{6}{l}{\emph{Robust}} \\
Wipe Board & Adam-U & $55.0$ & $70.0$ & $45.0$ & $60.0$ \\
Drawer Organization & FR3 & $50.0$ & $60.0$ & $40.0$ & $55.0$ \\
Wipe Board & FR3 & $45.0$ & $55.0$ & $35.0$ & $55.0$ \\
\rowcolor{gray!12} \multicolumn{2}{l}{\emph{Suite average}} & $50.0$ & $61.7$ & $40.0$ & $56.7$ \\
\midrule
\rowcolor{gray!25} \multicolumn{2}{l}{\textbf{Overall average}} & $\mathbf{60.8}$ & $\mathbf{67.1}$ & $\mathbf{50.0}$ & $\mathbf{63.8}$ \\
\bottomrule
\end{tabular}
}
\end{table}

\begin{figure}[p]
\centering
    \vspace{-2cm}
    \includegraphics[width=.81\linewidth]{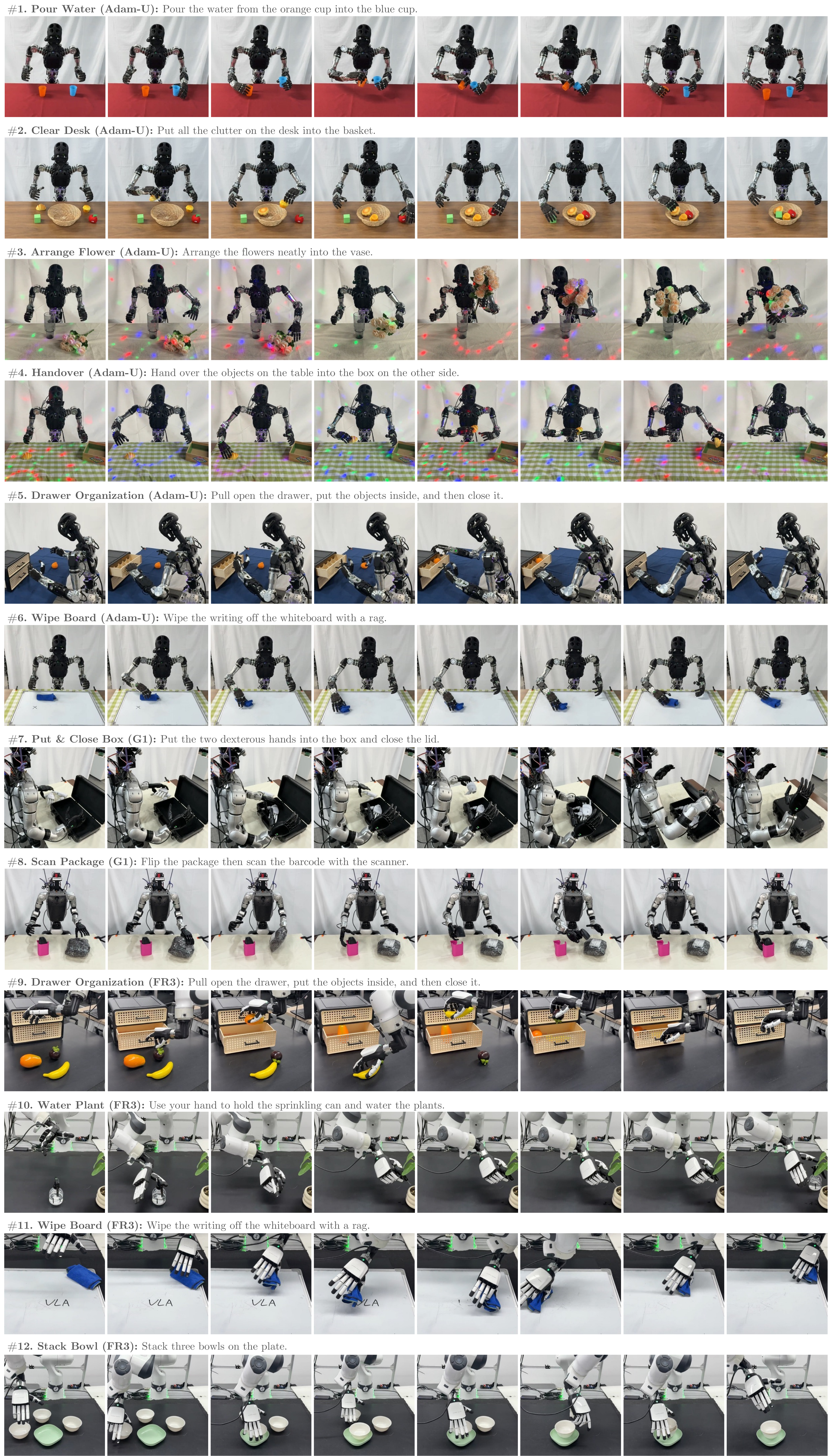}
    \caption{PTR rollouts on all $12$ real-robot tasks.}
    \label{fig:real-rollouts}
\end{figure}

\textbf{Task details and qualitative rollouts.}
Figure~\ref{fig:real-rollouts} shows example PTR rollouts for all $12$ official real-robot tasks.
The platform-specific task names are: Adam-U -- {\emph{Pour Water}, \emph{Clear Desk}, \emph{Arrange Flower}, \emph{Handover}, \emph{Drawer Organization}, \emph{Wipe Board}}; G1 -- {\emph{Put \& Close Box}, \emph{Scan Package}}; FR3 -- {\emph{Drawer Organization}, \emph{Water Plant}, \emph{Wipe Board}, \emph{Stack Bowl}}.
Each row displays eight keyframes from a single successful trial under the generalist PTR checkpoint, with tasks grouped by platform.

The Bimanual suite tests dual-hand coordination.
Pour Water on Adam-U requires one hand to hold the cup steady while the other tilts and pours, demanding precise bimanual force coordination to avoid spilling.
Handover on Adam-U demands coordinated timing and grip force during object transfer, where the object must be securely grasped before the transfer hand releases.
Put \& Close Box on G1 requires both dexterous hands to place objects into a box and then close the lid, a sequence where the post-action observation changes substantially between the placing and closing stages.

The Long-Horizon suite chains multiple subgoals.
Clear Desk on Adam-U involves sequential grasp-transport-place for all clutter objects, where the visual scene changes after each object removal and the chunk-level post-action signal can be harder to interpret than in single-stage placement tasks.
Drawer Organization on Adam-U requires pulling open a drawer, placing objects inside, and closing it, a sequence where the post-action observation changes dramatically at each stage.
Scan Package on G1 involves flipping a package to expose its barcode and then scanning it, requiring precise reorientation under dexterous hand control.

The Spatial suite tests placement accuracy.
Arrange Flower on Adam-U demands accurate insertion of flowers into a narrow vase aperture, where even small positional errors cause the stems to miss.
Stack Bowl on FR3 tests vertical alignment accuracy when stacking bowls onto a plate; the motion is relatively stereotyped across demonstrations, leaving less room for large gains from reweighting than in the more variable contact-rich tasks.
Water Plant on FR3 requires grasping a sprinkling can and tilting it at the correct angle over the plant, where the pour trajectory must be spatially precise.

The Robust suite randomizes scene conditions.
Wipe Board on Adam-U varies writing patterns and rag starting positions across resets; PTR shows one of the largest specialist per-task improvements here ($+15$\% over SFT), consistent with its tendency to avoid over-emphasizing training chunks whose post-action consequences are ambiguous due to visual variability.
Drawer Organization on FR3 shares the same semantic goal as its Adam-U counterpart but uses a different embodiment with different kinematics and camera viewpoints, testing cross-embodiment robustness directly.
Wipe Board on FR3 similarly mirrors the Adam-U wiping task under a different morphology.

Two task titles appear on both Adam-U and FR3 (Wipe Board and Drawer Organization), directly testing cross-embodiment robustness under shared semantics.
The semantics are shared but the raw trajectories are not, which is the regime PTR is explicitly designed to target.
Suite membership follows the official evaluation protocol rather than task title alone, which is why semantically related tasks can contribute to different suites across embodiments.

\begin{figure}[ht]
\centering
    \includegraphics[width=\linewidth]{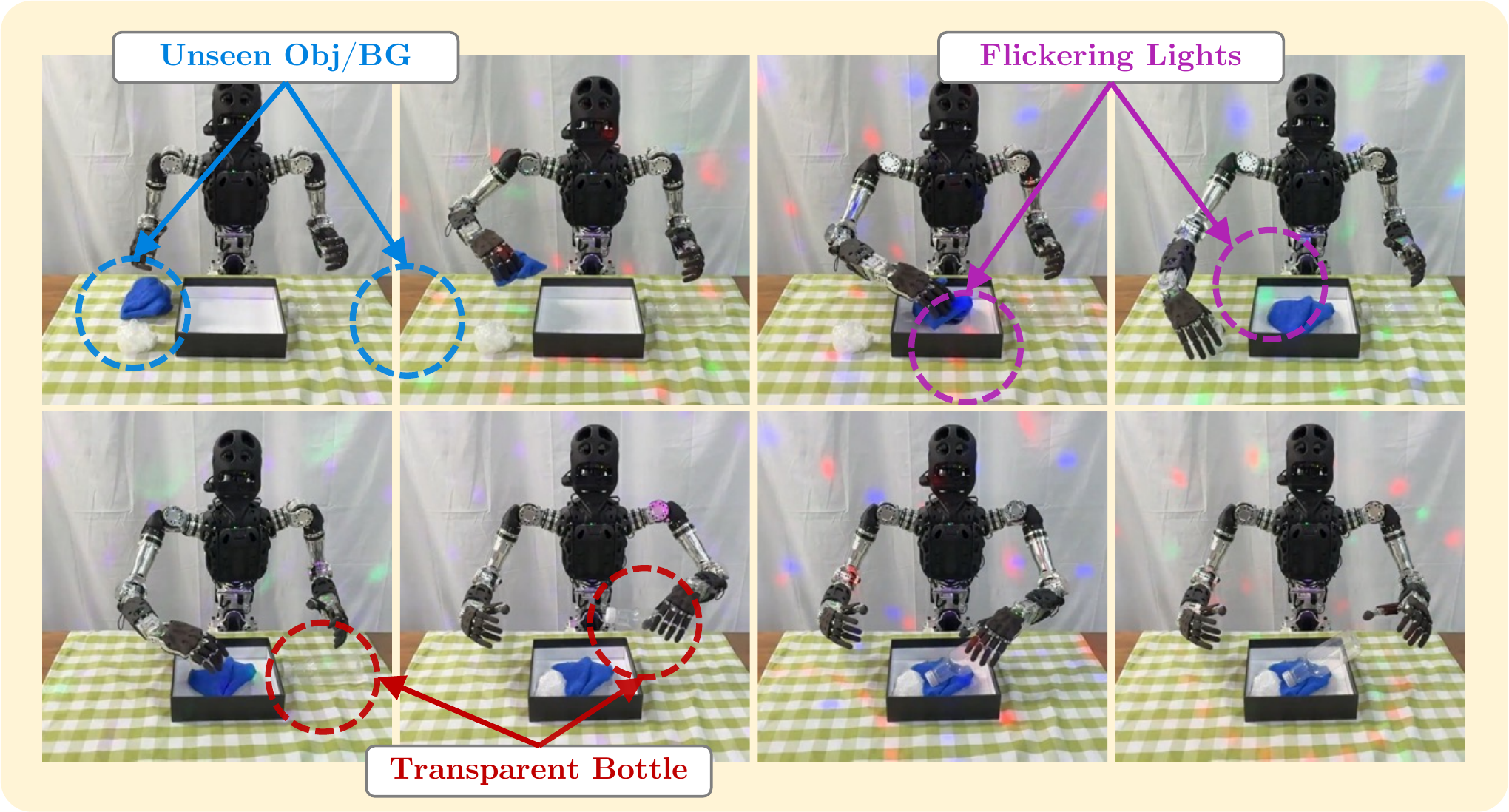}
    \caption{Illustrative hard out-of-distribution (OOD) evaluation with PTR. The scene contains unseen objects, a novel background, flickering overhead lighting, and a transparent bottle that challenges depth perception.}
    \label{fig:hard-ood-example}
\end{figure}

Figure~\ref{fig:hard-ood-example} presents a deliberately challenging OOD scene beyond the standard evaluation protocol, combining unseen objects and backgrounds, flickering overhead lighting, and a transparent bottle.
The PTR rollout still completes the intended task sequence.
This behavior is consistent with PTR's design: samples with ambiguous or noisy post-action consequences receive less relative emphasis during training, so learning is biased toward demonstrations where the action-to-outcome link is clearer.

\subsection{Ablations}
\label{sec:exp_ablation}

Table~\ref{tab:ablation} isolates the contribution of each PTR component on the standard (uncorrupted) benchmarks.
Weight clipping is the most critical component: removing it drops LIBERO by $2.3$ pp and RoboCasa by $6.5$ pp, confirming that bounding extreme weights is essential for stability.
The EMA target encoder ranks second ($-2.5$/$-5.4$ pp), because a frozen target space cannot track the evolving policy representation.
The adaptive scale controller contributes $0.7$/$3.7$ pp by adjusting $\tau_{\mathrm{score}}$ and $\beta$ online (Section~\ref{sec:adarefiner}).
Cross-rank negative gathering and belief proxy tokens each provide moderate gains ($0.8$/$1.3$ pp and $0.4$/$1.6$ pp) by enriching candidate-pool diversity and pre-action context quality.
Setting $\alpha{=}0$ recovers SFT+Belief, which trails PTR by $0.3$/$1.3$ pp, consistent with Table~\ref{tab:standard}.

The ablations separate three roles.
Clipping and $\beta$ control the \emph{scale} of the gradient reallocation.
The EMA target encoder keeps the scorer's target space aligned with the evolving backbone.
Belief tokens and richer negative pools improve the \emph{quality} of the identification problem itself.
The strong drops from removing clipping or EMA are consistent with PTR's design: without bounded weights or a stable target space, the score-to-weight map becomes either stale or overly volatile.

\begin{table}[t]
\centering
\caption{Ablation study on standard benchmarks (success \%). \textbf{Bold}: best; \underline{Underlined}: second best.}
\label{tab:ablation}
\setlength{\tabcolsep}{3.5pt}
\begin{tabular}{l cccc c ccc c}
\toprule
& \multicolumn{4}{c}{\textbf{LIBERO}} & & \multicolumn{3}{c}{\textbf{RoboCasa}} & \\
\cmidrule(lr){2-5}\cmidrule(lr){7-9}
\textbf{Configuration} & Spatial & Object & Goal & Long & \textbf{Avg} & Pick\&Pl. & Door/Dr. & Others & \textbf{Avg} \\
\midrule
\rowcolor{bestgreen!40}
PTR (full) & \textbf{98.0} & \textbf{98.8} & 97.4 & \textbf{97.0} & \textbf{97.8} & \textbf{38.3} & \textbf{73.0} & \underline{55.5} & \textbf{55.6} \\
\quad w/o belief tokens & 97.6 & 98.4 & \underline{97.6} & 95.8 & 97.4 & 36.8 & 71.0 & 54.2 & 54.0 \\
\quad w/o cross-rank gather & 97.2 & 98.2 & 97.0 & 95.4 & 97.0 & 35.7 & \underline{71.7} & \textbf{55.5} & 54.3 \\
\quad w/o EMA (frozen enc.) & 95.8 & 96.6 & 95.2 & 93.4 & 95.3 & 33.0 & 67.3 & 50.5 & 50.2 \\
\quad w/o refiner & \underline{97.8} & 97.8 & 96.4 & \underline{96.2} & 97.1 & 34.7 & 69.3 & 52.0 & 51.9 \\
\quad w/o clipping & 96.2 & 97.0 & 95.6 & 93.0 & 95.5 & 32.3 & 66.0 & 49.3 & 49.1 \\
\quad $\alpha{=}0$ (SFT+Belief) & 97.4 & \underline{98.6} & \textbf{98.0} & 96.0 & \underline{97.5} & \underline{36.7} & 71.5 & 55.0 & \underline{54.3} \\
\bottomrule
\end{tabular}
\end{table}

\subsection{Training analysis}
\label{sec:exp_training}

\begin{figure}[ht]
    \centering
    \includegraphics[width=.95\linewidth]{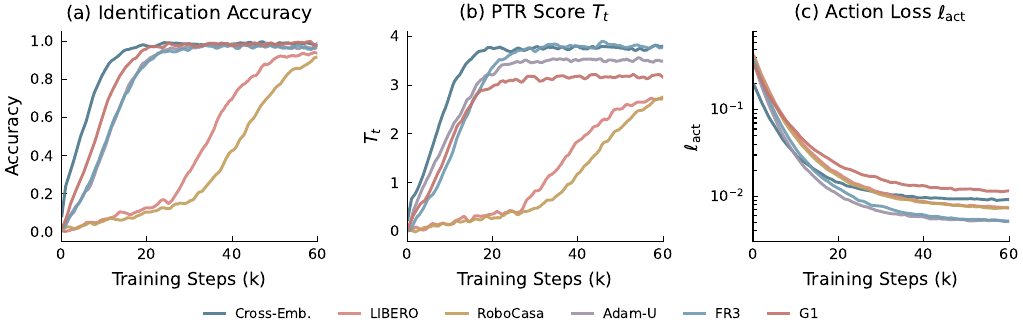}
    \caption{PTR training dynamics across six embodiment settings ($60$k steps). (a)~Identification accuracy shows two convergence groups; the legend applies to all three subplots. (b)~PTR score $T_t$ stabilizes at embodiment-dependent levels. (c)~Action loss $\ell_{\mathrm{act}}$ decreases steadily.}
    \label{fig:training_analysis}
\end{figure}

\textbf{Training dynamics.}
Figure~\ref{fig:training_analysis} tracks three core PTR quantities across six embodiment settings during $60$k training steps.
Two convergence groups emerge in identification accuracy~(a): Cross-Emb., Adam-U, FR3, and G1 improve markedly earlier, while LIBERO and RoboCasa rise more gradually and reach their plateau later in training.

The PTR score $T_t$~(b) shows a similar pattern: the earlier-converging settings stabilize around $3.2$--$3.8$, while the slower group reaches roughly $2.8$--$3.1$.
The cross-embodiment joint run achieves the highest final score (${\sim}3.8$), consistent with the richer candidate pool from five data sources.
The action loss $\ell_{\mathrm{act}}$~(c) decreases by over an order of magnitude for all settings, from $0.2$--$0.4$ down to $0.005$--$0.012$.

\begin{figure}[t]
    \centering
    \includegraphics[width=.7\linewidth]{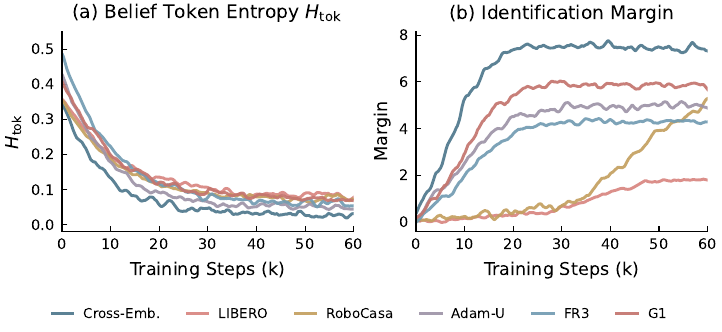}
    \caption{Belief token and identification margin dynamics. (a)~Belief token entropy $H_{\mathrm{tok}}$ converges to near-zero across all settings. (b)~Identification margin grows steadily, with cross-embodiment training achieving the highest discriminability.}
    \label{fig:belief_nce_margin}
\end{figure}

\textbf{Belief tokens and identification margin.}
Figure~\ref{fig:belief_nce_margin} provides two complementary views.
Belief token entropy $H_{\mathrm{tok}}$~(a) drops steadily from roughly $0.3$--$0.5$ toward a low plateau below $0.1$, indicating that the soft causal tokenizer converges to compact slot assignments regardless of embodiment.
The identification margin~(b), the gap between the matched-target logit and the hardest negative, grows from near-zero values to approximately $1.9$--$7.5$ depending on the setting.
The cross-embodiment run achieves the highest final margin (${\sim}7.5$), consistent with diverse data providing more distinctive post-action signatures.
LIBERO exhibits the lowest margin (${\sim}1.9$) despite reasonable accuracy, suggesting that its homogeneous task structure produces less separable target embeddings.
Accuracy saturates once the positive is usually top-1 in the candidate set, whereas the margin can continue to grow as hard negatives become better separated.

\begin{figure}[t]
    \centering
    \includegraphics[width=\linewidth]{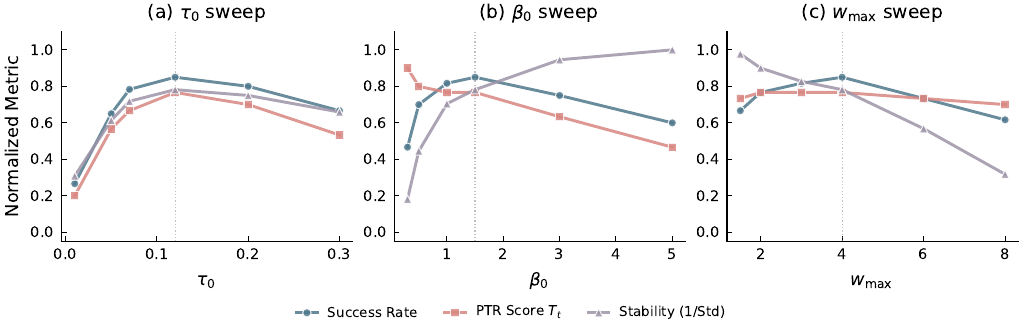}
    \caption{Hyperparameter sensitivity on cross-embodiment LIBERO. Each subplot sweeps one parameter with others fixed at defaults, tracking three normalized diagnostics: success rate, final PTR score $T_t$, and weight stability $1/\mathrm{Std}(w)$. Dashed lines mark the default values. Legend is shared across all three subplots.}
    \label{fig:hyperparam}
\end{figure}

\textbf{Hyperparameter sensitivity.}
Figure~\ref{fig:hyperparam} sweeps the three key hyperparameters ($\tau_0$, $\beta_0$, $w_{\max}$) on the cross-embodiment LIBERO setting.
The scorer temperature $\tau_0$~(a) has a clear sweet spot near $0.12$: smaller values make the posterior too sharp and reduce stability, while larger values over-flatten the posterior and weaken the score signal.
The advantage scaling $\beta_0$~(b) controls weight spread directly: very small values amplify score differences into unstable weights, while very large values compress weights toward uniformity and effectively disable reweighting.
The clipping bound $w_{\max}$~(c) provides a complementary safety mechanism: $w_{\max}=4.0$ gives the best balance in this sweep, whereas larger values allow less controlled gradients from extreme weights.
The default configuration ($\tau_0{=}0.12$, $\beta_0{=}1.5$, $w_{\max}{=}4.0$) achieves the best overall balance across the three sweeps shown here.

\begin{figure}[ht]
    \centering
    \includegraphics[width=\linewidth]{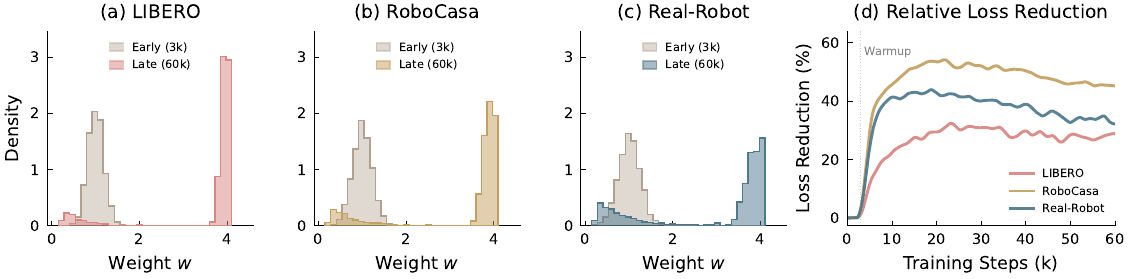}
    \caption{PTR weight evolution across benchmarks. (a)--(c)~Sample weight distribution at early stage ($3$k steps, grey) vs.\ late stage ($60$k steps, colored) for LIBERO, RoboCasa, and real-robot generalist training. (d)~Relative loss reduction $(\ell_{\mathrm{SFT}} - \ell_{\mathrm{PTR}})/\ell_{\mathrm{SFT}}$ over training.}
    \label{fig:ptr_vs_sft}
\end{figure}

\textbf{Weight evolution and loss reduction.}
Figure~\ref{fig:ptr_vs_sft} visualizes how PTR's sample weights evolve during training.
Subplots~(a)--(c) overlay the weight distribution at two snapshots: early stage ($3$k steps, grey) when the NCE scorer has just finished warmup and weights remain near-uniform around $w{=}1$, versus late stage ($60$k steps, colored) after the scorer has fully matured.

LIBERO~(a), with relatively homogeneous simulation data, concentrates much of its mass near the clipping bound $w_{\max}{=}4.0$ and retains only a thin suppressed tail near $w_{\min}$.

RoboCasa Human \emph{50 shots}~(b), with more diverse scenes and noisier demonstrations, shows a visibly broader suppressed region.

The real-robot generalist setting~(c) exhibits the widest spread, reflecting cross-embodiment heterogeneity where kinematically mismatched demonstrations are less likely to receive large identification scores.

Subplot~(d) shows the relative loss reduction $(\ell_{\mathrm{SFT}} - \ell_{\mathrm{PTR}})/\ell_{\mathrm{SFT}}$ over training.
All three benchmarks show near-zero reduction during the $3000$-step warmup and then diverge as the scorer activates.
RoboCasa reaches the largest late-stage reduction, consistent with a noisier data distribution where PTR has more room to suppress uninformative samples.
The real-robot generalist setting shows earlier onset of improvement, which is consistent with the richer candidate pool from cross-embodiment data.

\textbf{Adaptive controller dynamics over training.}
The refiner's controller signals reveal a consistent two-phase pattern.
During the first ${\sim}5$k steps (including the $3000$-step NCE warmup), $\tau_{\mathrm{score}}$ stays near its initial value of $0.12$ and the hard-negative ratio remains at $0.0$.
Once the identification head begins producing meaningful gradients, $\tau_{\mathrm{score}}$ decreases steadily to approximately $0.03$, reflecting a sharpening posterior.
Concurrently, the hard-negative ratio ramps from $0.0$ to approximately $0.50$, and $\beta$ shows a mild downward trend from $1.5$ to approximately $1.2$, consistent with the refiner allowing stronger weight differentiation as the scorer matures.

\textbf{Ablation training curves.}
The variants without clipping (w/o clip) and without EMA (w/o EMA) exhibit noticeably less stable dynamics.
Without clipping, the weight distribution develops heavy tails early in training: a small fraction of samples receive weights above $10$, causing gradient spikes that manifest as periodic loss oscillations.
Without EMA, the target space drifts with backbone updates, creating a moving-target problem for the scorer; identification accuracy fluctuates rather than converging monotonically, and the resulting weight signal is noisy.
Both failure modes are consistent with the design rationale in Section~\ref{sec:ptr_reweight}.

\section{Conclusion and Limitations}
\label{sec:concl}

PTR is a conservative offline post-training method that reallocates credit across heterogeneous robot demonstrations without reward labels or a tractable policy likelihood.
By scoring each sample through a candidate-set identification posterior over post-action consequences, PTR provides two complementary benefits: a floor that removes extra emphasis from ambiguous or noisy demonstrations and suppresses clearly counter-evidential ones, and a ceiling that amplifies useful cross-embodiment transfer when pooled data cover task-relevant regions.

The exponential weight mapping, clipping, and self-normalization keep the induced training distribution close to the original data, making the method conservative by construction.

Three limitations remain.
PTR is most informative when post-action observations are available; chunks without a usable future observation revert to uniform weighting, and purely real-time streaming scenarios remain outside the method's scope.
The identification signal depends on the quality of the learned representation: a poorly pretrained backbone limits the scorer's discriminability, and PTR falls back to near-uniform weighting.
Finally, PTR improves the effective training distribution but does not directly optimize task success; it is a data-curation mechanism rather than a policy-optimization algorithm.

\clearpage
\bibliographystyle{unsrtnat}
\bibliography{ref}

\clearpage
\appendix

\section{Proofs}
\label{app:proofs}

This appendix collects the formal proofs of the three propositions stated in the main text.

\subsection{Proof of Proposition~\ref{prop:density_ratio} (Density-ratio form)}

\begin{proof}
Let $p_+(y):=p(y\mid h,e)$ and $p_-(y):=p_N(y\mid h)$.
Under the symmetric candidate-set model from Section~\ref{sec:candidate_model}, conditioned on $(h,e)$ and the full candidate set $Y=(Y_0,\ldots,Y_K)$, the Bayes posterior for candidate position $j$ is
\begin{align}
\label{eq:app_ratio_step1}
p^\star(I=j\mid h,e,Y)
&=
\frac{p_+(Y_j)\prod_{\ell\neq j}p_-(Y_\ell)}{\sum_{m=0}^{K} p_+(Y_m)\prod_{\ell\neq m}p_-(Y_\ell)}
\notag\\[4pt]
&=
\frac{\frac{p_+(Y_j)}{p_-(Y_j)}}{\sum_{m=0}^{K}\frac{p_+(Y_m)}{p_-(Y_m)}},
\end{align}
where the second line divides numerator and denominator by $\prod_{\ell=0}^{K} p_-(Y_\ell)$.
Taking posterior odds between any two candidate positions $j$ and $m$ gives
\begin{equation}
\label{eq:app_logodds}
\log\frac{p^\star(I=j\mid h,e,Y)}{p^\star(I=m\mid h,e,Y)}
=
\log\frac{p_+(Y_j)}{p_-(Y_j)}
-
\log\frac{p_+(Y_m)}{p_-(Y_m)}.
\end{equation}
A $(K{+}1)$-way softmax classifier represents exactly such pairwise log-odds via differences of logits.
Therefore the shared scoring function must agree with $\log r(y)$ up to a candidate-independent additive constant: fixing any reference value $y_{\mathrm{ref}}$ on the common support gives
$s^\star(h,e,y)-s^\star(h,e,y_{\mathrm{ref}})=\log r(y)-\log r(y_{\mathrm{ref}})$,
so $s^\star(h,e,y)=\log\frac{p_+(y)}{p_-(y)}+b(h,e)$ for some $b(h,e)$ independent of $y$.
Substituting back $p_+(y)=p(y\mid h,e)$ and $p_-(y)=p_N(y\mid h)$ yields Eq~\eqref{eq:app_density_ratio}.
\end{proof}

\subsection{Proof of Proposition~\ref{prop:kl_lens_main} (KL lens)}

\begin{proof}[Proof of Proposition~\ref{prop:kl_lens_main} under the bounded-ratio regularity condition]
By Proposition~\ref{prop:density_ratio}, Bayes-optimal shared per-candidate logits equal $\log r(y)$ up to an additive constant that cancels in the softmax.
The corresponding identification posterior therefore takes the form
\begin{equation}
\label{eq:app_opt_post}
p^\star(I=0\mid h,e,Y)
=
\frac{r(Y_0)}{\sum_{j=0}^{K} r(Y_j)}.
\end{equation}
Define the empirical average
\[
A_K := \frac{1}{K{+}1}\sum_{j=0}^{K} r(Y_j).
\]
Substituting Eq~\eqref{eq:app_opt_post} into the definition of the PTR score gives
\begin{align}
\label{eq:app_score_expand}
T^\star
&=
\log\frac{p^\star(I=0\mid h,e,Y)}{1/(K{+}1)}
\notag\\
&=
\log r(Y_0)-\log A_K.
\end{align}
Because each negative target $Y_j$ ($j\ge 1$) is drawn from $p_-$, the density ratio has baseline mean one:
\begin{align}
\E_{Y\sim p_-}[r(Y)]
&=
\int \frac{p_+(y)}{p_-(y)}\,p_-(y)\,dy
\notag\\
&=
\int p_+(y)\,dy
\;=\;
1.
\end{align}
By the strong law of large numbers, the empirical average of the negative density ratios converges almost surely:
\begin{equation}
\label{eq:app_lln}
\frac{1}{K}\sum_{j=1}^{K} r(Y_j)\ \longrightarrow\ 1
\quad\text{almost surely as }K\to\infty.
\end{equation}
The single matched term contributes at most $C/(K{+}1)$, so $A_K\to 1$ almost surely; equivalently,
\begin{equation}
\label{eq:app_den_limit}
\frac{1}{K{+}1}\sum_{j=0}^{K} r(Y_j)\ \longrightarrow\ 1
\quad\text{almost surely}.
\end{equation}
Plugging Eq~\eqref{eq:app_den_limit} into Eq~\eqref{eq:app_score_expand} yields the pointwise limit
\begin{equation}
\label{eq:app_score_pointwise}
T^\star\ \longrightarrow\ \log r(Y_0)
\quad\text{almost surely}.
\end{equation}
Under $0<c\le r(y)\le C<\infty$, we have $A_K\in[c,C]$ for every $K$, hence $|T^\star|\le \log(C/c)$.
Dominated convergence therefore applies, and taking expectations over $Y_0\sim p_+$ gives
\begin{align}
\E[T^\star\mid h,e]
&\ \longrightarrow\
\E_{Y_0\sim p_+}[\log r(Y_0)]
\notag\\
&=
\int p_+(y)\log\frac{p_+(y)}{p_-(y)}\,dy
\;=\;
\KL(p_+\|p_-),
\end{align}
which is Eq~\eqref{eq:kl_score_main}.
\end{proof}

\subsection{Proof of Proposition~\ref{prop:mixture_main} (Source reweighting)}

\begin{proof}[Proof of Proposition~\ref{prop:mixture_main}]
Assume the dataset is a mixture
\begin{equation}
p_{\cD}(x)=\sum_m \pi_m p_m(x),
\end{equation}
where $m$ indexes sources (embodiments, operators, or domains) and $\pi_m$ is the prior proportion of source $m$.
Assume also that $\E_{p_m}[\exp(J(x)/\beta)]<\infty$ for every source $m$, so the tilted source marginals are well-defined.
Formally, let $M$ be an observed source label and write the joint distribution as $p_{\cD}(x,m)=\pi_m p_m(x)$.
Tilting by the score $J(x)$ with temperature $\beta$ produces the joint induced distribution $q^\star(x,m)\propto \pi_m p_m(x)\exp(J(x)/\beta)$, whose source marginal is $q^\star(m)=\int q^\star(x,m)\,dx$.
Let $Z$ denote the normalizing constant.
Using the joint form $q^\star(x,m)=\frac{1}{Z}\pi_m p_m(x)\exp(J(x)/\beta)$, the source marginal is
\begin{align}
q^\star(m)
&=
\int q^\star(x,m)\,dx
\notag\\[4pt]
&=
\frac{1}{Z}\int \pi_m p_m(x)\exp\!\bigl(J(x)/\beta\bigr)\,dx.
\end{align}
Pulling out $\pi_m$ and rewriting the integral as an expectation gives
\begin{align}
q^\star(m)
&=
\frac{\pi_m}{Z}\int p_m(x)\exp\!\bigl(J(x)/\beta\bigr)\,dx
\notag\\[4pt]
&=
\frac{\pi_m\,\E_{p_m}\!\bigl[\exp(J(x)/\beta)\bigr]}{Z}.
\end{align}
By construction, $Z=\sum_j \pi_j\,\E_{p_j}[\exp(J(x)/\beta)]$.
Substituting this expression for $Z$ yields the closed-form expression
\begin{equation}
q^\star(m)
=
\frac{\pi_m\,\E_{p_m}\!\bigl[\exp(J(x)/\beta)\bigr]}
     {\sum_{j} \pi_j\,\E_{p_j}\!\bigl[\exp(J(x)/\beta)\bigr]},
\end{equation}
which is Eq~\eqref{eq:mixture_main}.
\end{proof}

\clearpage
\section{Training Configuration}
\label{app:hyperparams}

Table~\ref{tab:hyperparams} lists the complete hyperparameter configuration used for all PTR experiments.
All experiments use these defaults unless a parameter is explicitly varied in a sensitivity sweep (Section~\ref{sec:exp_training}).

\begin{table}[ht]
\centering
\caption{Complete hyperparameter listing for PTR post-training.}
\label{tab:hyperparams}
\small
\setlength{\tabcolsep}{4pt}
\resizebox{.73\linewidth}{!}{
\begin{tabular}{l l l}
\toprule
\textbf{Parameter} & \textbf{Symbol} & \textbf{Value} \\
\midrule
\multicolumn{3}{l}{\emph{Optimization}} \\
Optimizer & -- & AdamW \\
Learning rate & $\eta$ & $10^{-4}$ \\
Weight decay & -- & $0.01$ \\
Schedule & -- & Cosine with $2000$-step linear warmup \\
Total steps & -- & $60$k \\
Global batch size & -- & $128$ \\
\midrule
\multicolumn{3}{l}{\emph{Base policy}} \\
Unified action dimension & -- & $200$ \\
Action chunk length & $L$ & $16$ \\
Flow-matching time-step sampling & -- & $\mathrm{Beta}(1.5,\,1.0)$ \\
Inference denoising steps & -- & $4$ \\
\midrule
\multicolumn{3}{l}{\emph{PTR scorer}} \\
Identification loss weight & $\lambda_{\mathrm{id}}$ & $0.05$ \\
NCE warmup steps & -- & $3000$ \\
Scorer temperature (init) & $\tau_{\mathrm{score}}$ & $0.12$ \\
Advantage scaling (init) & $\beta$ & $1.5$ \\
Logit clamp & -- & $20.0$ \\
Clipping lower bound & $w_{\min}$ & $0.25$ \\
Clipping upper bound & $w_{\max}$ & $4.0$ \\
Mixture coefficient & $\alpha$ & $1$ \\
\midrule
\multicolumn{3}{l}{\emph{Candidate pool}} \\
FIFO queue size & -- & $1024$ \\
Max queue negatives per sample & -- & $64$ \\
Cross-rank gather & -- & Enabled ($8$ GPUs) \\
\midrule
\multicolumn{3}{l}{\emph{EMA target encoder}} \\
EMA decay & $\mu$ & $0.999$ \\
Source layer & -- & Layer $12$ of InternViT-300M \\
L2 normalization eps & -- & $10^{-6}$ \\
\midrule
\multicolumn{3}{l}{\emph{Belief tokenizer}} \\
Number of belief tokens & $M$ & $4$ \\
Tokenizer temperature & $\tau_{\mathrm{tok}}$ & $1.0$ \\
Entropy regularizer & $\lambda_{\mathrm{ent}}$ & $10^{-3}$ \\
Diversity regularizer & $\lambda_{\mathrm{div}}$ & $10^{-3}$ \\
\midrule
\multicolumn{3}{l}{\emph{Action sensitivity}} \\
Ranking loss weight & $\lambda_{\mathrm{rank}}$ & $0.25$ \\
\midrule
\multicolumn{3}{l}{\emph{Adaptive scale control}} \\
EMA momentum & -- & $0.98$ \\
$\tau_{\mathrm{score}}$ bounds & $[\tau_{\min},\tau_{\max}]$ & $[0.03,\,0.20]$ \\
$\beta$ bounds & $[\beta_{\min},\beta_{\max}]$ & $[0.5,\,3.0]$ \\
Hard-negative ratio bounds & -- & $[0.0,\,0.5]$ \\
\bottomrule
\end{tabular}
}
\end{table}

\end{document}